\documentclass{article}

\usepackage[final,main]{neurips_2026}
\usepackage{graphicx} 

\usepackage{longtable}

\usepackage[utf8]{inputenc}
\usepackage[T1]{fontenc}
\usepackage{hyperref}
\usepackage{url}
\usepackage{booktabs}
\usepackage{amsfonts}
\usepackage{nicefrac}
\usepackage{microtype}
\usepackage{xcolor}
\usepackage{amsmath, amssymb, amsthm}
\usepackage{graphicx}
\usepackage{subcaption}
\usepackage{multirow}
\usepackage{enumitem}
\usepackage{array}
\usepackage{wrapfig}
\usepackage{mathtools}
\usepackage{bbm}
\usepackage{float}
\usepackage{longtable}
\usepackage{booktabs}
\usepackage{xcolor}

\newcommand{\stat}[2]{%
  \ensuremath{#1\,\scalebox{0.8}{$\pm #2$}}%
}

\usepackage[ruled,vlined,linesnumbered]{algorithm2e}
\DontPrintSemicolon
\SetKwInOut{KwIn}{Input}
\SetKwInOut{KwOut}{Output}

\newcommand{\moesvd}{MoE-SVD}
\newcommand{\naee}{NAEE}
\newcommand{\moeiitwo}{MoE-I$^2$}

\definecolor{citecolor}{HTML}{0071bc}
\hypersetup{
    colorlinks=true,
    linkcolor=red,
    citecolor=citecolor,
    filecolor=magenta,      
    urlcolor=magenta,
}

\title{FlexMoE: Learning One-for-All Multi-Budget Expert Pruning for Flexible MoE Deployment}

\author{%
\begin{tabular}{@{}c@{\hspace{3em}}c@{\hspace{2em}}c@{}}
\shortstack[c]{
  Fan Mo$^{1,*}$ \\
  {\footnotesize\texttt{fan.mo@u.nus.edu}}
}
&
\shortstack[c]{
  Yuxuan Han$^{1,*}$ \\
  {\footnotesize\texttt{han\_yuxuan@u.nus.edu}}
}
&
\shortstack[c]{
  Geng Zhang$^{1,*}$ \\
  {\footnotesize\texttt{zhangg@comp.nus.edu.sg}}
}
\\[1.5em]
\multicolumn{3}{c}{%
  \shortstack[c]{
    Wangbo Zhao$^{2,\dagger}$ \\
    {\footnotesize\texttt{wangbo.zhao96@gmail.com}}
  }
  \hspace{2em}
  \shortstack[c]{
    Yang You$^{1,\dagger}$ \\
    {\footnotesize\texttt{yangyou@nus.edu.sg}}
  }
}
\\[2em]
\multicolumn{3}{c}{
  $^{1}$ National University of Singapore
}
\\[0.25em]
\multicolumn{3}{c}{
  $^{2}$ The Hong Kong University of Science and Technology
}
\\[1em]
\multicolumn{3}{c}{
  {\small $^{*}$ Equal contribution.
  \quad $^{\dagger}$ Corresponding authors.}
}
\end{tabular}%
}

\begin{document}

\maketitle
\begin{abstract}
\label{sec:abs}

Mixture-of-Experts (MoE) language models scale model capacity through sparse expert activation, yet deployment still requires storing and serving all experts. Moreover, deployment budgets can vary substantially across devices, users, and workloads. Existing MoE compression methods typically optimize a separate compressed model for each target budget. We introduce FlexMoE, a post-training one-for-all framework that converts a large pretrained MoE model into a multi-budget family of smaller deployable subnetworks. FlexMoE first ranks expert FFN channels by importance to construct a sliceable weight space shared across budgets, and then learns expert-wise discrete retention actions under gradually increasing cost pressure, allowing a single action-training run to produce reliable pruning masks across budgets. Moreover, only a single mid-budget fine-tune can further recover degraded performance and yields a shared recovered model that transfers to other unseen budgets, avoiding repeated per-budget recovery and separated checkpoint storage. Across tested MoE LLMs with diverse architectures and scales, FlexMoE consistently outperforms strong MoE compression baselines. In particular, on Qwen2-57B-A14B, it retains ~99.8\% of base performance while pruning 50\% of routed-expert parameters even without recovery fine-tuning. For deployment, pruned subnetworks deliver real memory reduction and end-to-end throughput gains, while our kernel co-design enables runtime budget switching without checkpoint reloading and delivers substantial inference acceleration.
\end{abstract}

\section{Introduction}
\label{sec:intro}

Mixture-of-Experts (MoE) models scale parameter capacity while activating only a small subset of experts per token, achieving strong quality with substantially lower active computation \citep{lepikhin2021gshard,fedus2022switch}. This sparse-expert design has therefore become an increasingly common recipe for scaling modern foundation models.

However, sparse activation does not eliminate the deployment cost of storing and serving all experts. In practice, the available deployment budget---including memory, latency/throughput requirements, context demands, and hardware capacity---can vary substantially across users, devices, and workloads. On deployment side, one service may need long-context or agent-level reasoning under a generous budget, while another may serve lightweight chatbot requests under much stricter latency and cost constraints \citep{fu2024amoeballm,gao2026flyingserving,ma2026orbitflow}. MoE deployment and compression is therefore not only a model-reduction problem, but also a deployment-adaptation problem.

\begin{figure}[t]
    \centering
    \includegraphics[width=0.9\textwidth]{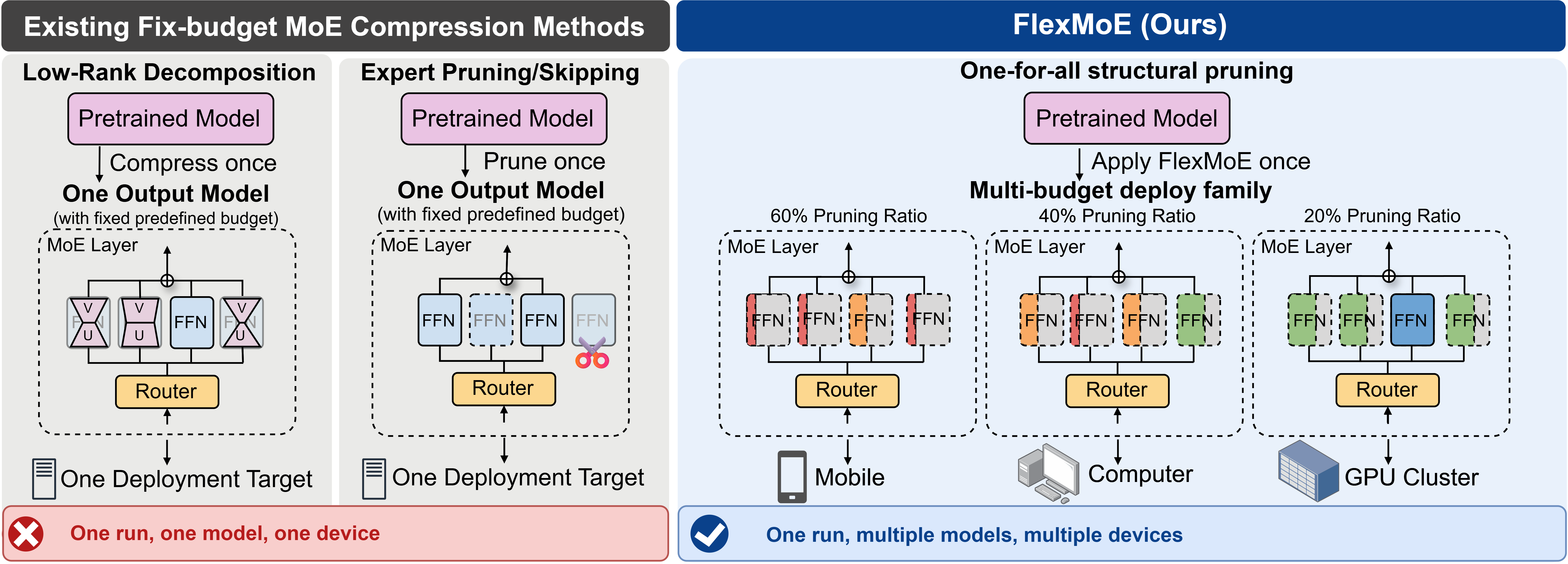}
    \caption{Existing MoE compression methods typically optimize a compressed endpoint for a target budget, whereas FlexMoE learns a shared multi-budget pruning family from one pretrained model, avoiding separate compression and checkpoints across deployment targets.}
    \label{fig:fixed_budget_vs_flexmoe}
\end{figure}

Existing MoE compression methods are nevertheless largely fixed-budget. Expert pruning, skipping, or merging and expert-internal compression typically optimize one compressed endpoint for a chosen target budget \citep{lu2024naee,bai2025diep,yang2024moei2,gu2025d2moe,li2025moesvd,xu2026tdmoe}. When the operating budget changes, one may need to rerun compression, reload a different model, or maintain multiple budget-specific checkpoints. This limitation becomes particularly relevant under non-stationary serving workloads, where runtime reconfiguration can be valuable \citep{gao2026flyingserving,qian2026kevlarflow}.

\begin{figure}
    \centering
    \includegraphics[width=\textwidth]{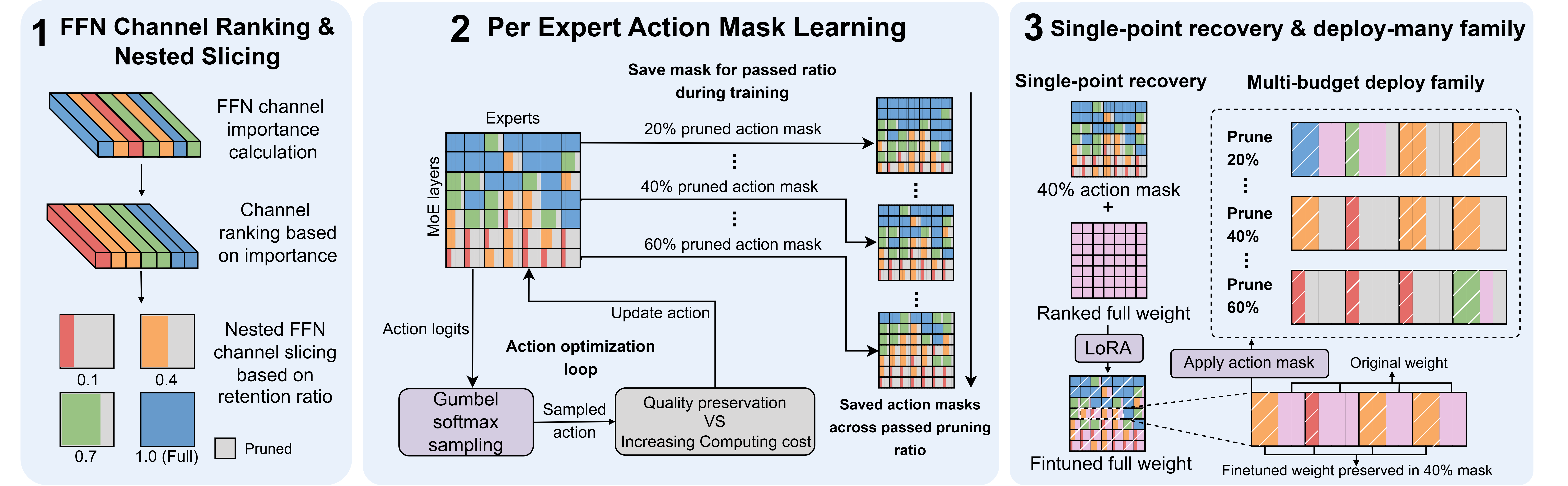}
    \caption{Overview of the FlexMoE pipeline. FlexMoE first importance-ranks expert FFN channels to construct a shared sliceable weight space across multiple retention ratios, then learns expert-wise retention (pruning) actions across pruning budgets. An optional single-point recovery fine-tune produces full recovered weights reusable across the resulting multi-budget subnet family.}
    \label{fig:method-pipeline}
\end{figure}

To address this, we propose \textbf{FlexMoE}, a post-training one-for-all intra-expert pruning framework that converts one pretrained MoE model into a family of materializable subnetworks, which are nested within the full model with original size and share its weights across multiple budgets. As illustrated in Figure~\ref{fig:method-pipeline}, FlexMoE first ranks expert FFN channels by importance to construct a shared sliceable weight space, and then learns expert-wise discrete retention actions under gradually increasing pruning pressure. A single action-training run thereby exports pruning masks across multiple budgets without saving actual model checkpoints. An optional single mid-budget recovery fine-tune further yields shared recovered weights reusable across the budget family. We additionally co-design the inference kernel to realize efficient online budget switching.

We evaluate FlexMoE on Mixtral-8x7B, Phi-3.5-MoE, and Qwen2-57B-A14B, where it consistently outperforms strong MoE compression baselines, and the single mid-budget recovery transfers well to unseen higher and lower budgets. In particular, on Qwen2-57B-A14B, the pruned subnet retains about 99.8\% of the base performance at 50\% pruning budget and still preserves about 92.9\% at 80\% budget. On deployment side, the resulting subnet family provides real end-to-end memory reduction, throughput gains, and enables runtime budget switching.


\section{Related Work}

\subsection{MoE Compression}

MoE compression broadly targets either expert-level redundancy or expert-internal weights. At the expert level, NAEE exploits unequal expert importance for expert pruning and skipping \citep{lu2024naee}; DiEP learns expert retention through differentiable pruning \citep{bai2025diep}; HC-SMoE clusters and merges functionally similar experts without retraining \citep{chen2025hcsmoe}; and MoNE replaces redundant expert outputs with lightweight novices\citep{zhang2026mone}. Another line compresses expert weights while largely preserving the original MoE structure: MoE-SVD and D$^2$-MoE perform expert-internal compression, TD-MoE further introduces cross-expert joint tensor decomposition within each layer, and MoE-I$^2$ combines inter-expert pruning with intra-expert decomposition \citep{li2025moesvd,gu2025d2moe,xu2026tdmoe,yang2024moei2}. FlexMoE is closest to this intra-expert family, but instead learns fine-grained prefix-slicing actions over a shared importance-ordered expert weight space, enabling multiple deployment budgets from the same pretrained MoE while preserving its router allocation and expert topology.

\subsection{Nested Subnetworks, Mask Learning, and Ranking}

The broader train-once, deploy-many (one-for-all) paradigm is related to elastic and nested subnetwork training. US-Nets introduced universally slimmable networks, while MatFormer, Flextron, and AmoebaLLM extended nested or many-in-one parameter sharing to transformers and large language models \citep{yu2019usnets,devvrit2024matformer,cai2024flextron,fu2024amoeballm}. More recent MoE methods introduce elasticity through coarse-to-fine expert ranking or slimmable expert widths \citep{wang2026mmoe,tastan2026mose}. In contrast, FlexMoE starts from an already pretrained MoE LLM and derives its multi-budget subnet family at the post-training stage.

FlexMoE's shared expert slicable weight space also builds on differentiable model structure learning and gradient-based importance ranking. MaskLLM and Gumbel-Softmax provide tools for gradient-based discrete structural decisions \citep{fang2024maskllm,jang2017gumbel}, while RECAP and earlier Taylor criteria use grouped first-order saliency for structured pruning \citep{ilhan2024recap,molchanov2019importance}. We combine these ingredients to learn per-expert prefix-slicing actions over importance-ordered FFN channels.


\section{Method}
\label{sec:method}

We consider a pretrained MoE LLM with $L$ MoE layers and $E$ experts per layer.
For a hidden state $h$ routed to expert $e$ in layer $l$, its FFN is
\begin{equation}
\mathrm{FFN}_{l,e}(h)
=
W^{\mathrm{down}}_{l,e}
\Big(
\phi(W^{\mathrm{gate}}_{l,e}h)
\odot
W^{\mathrm{up}}_{l,e}h
\Big),
\label{eq:expert_ffn}
\end{equation}
where $d_{\mathrm{ff}}$ denotes the intermediate FFN width.

\subsection{Intra-Expert Channel Importance Ranking}

Directly applying channel prefix slicing to pretrained experts is brittle. We therefore perform a one-time reordering step that converts each expert FFN into an importance-ordered layout, so that prefixes preserve more important hidden-channels under prefix-slice pruning.

For expert $(l,e)$, let $\Theta_{l,e,j}$ denote all parameters associated with hidden channel $j$: the corresponding rows in expert weights $W^{\mathrm{gate}}_{l,e}$ and $W^{\mathrm{up}}_{l,e}$ and the matching column of $W^{\mathrm{down}}_{l,e}$.
Since FFN hidden channels are permutation-invariant when the same permutation is applied consistently to these projections~\citep{navon2023equivariant}, they can be reordered without changing the calculation correctness and therefore the full expert function.

We rank channels using grouped first-order Taylor saliency~\citep{molchanov2019importance,ilhan2024recap}. 
On calibration sample batches $\mathcal B_{\mathrm{rank}}$, the importance score of channel $j$ is
\begin{equation}
s_{l,e,j}
=
\frac{1}{|\mathcal B_{\mathrm{rank}}|}
\sum_{b\in\mathcal B_{\mathrm{rank}}}
\sum_{\theta\in\Theta_{l,e,j}}
\left(
\theta
\frac{\partial \mathcal L^{(b)}}{\partial \theta}
\right)^2 .
\label{eq:importance_score}
\end{equation}
Intuitively, $s_{l,e,j}$ estimates the extent of param-group $\Theta_{l,e,j}$ (channel $j$) affects  the loss $\mathcal L$. Sorting $s_{l,e,j}$ in descending order gives permutation matrix $P_{l,e}$, which is applied consistently across the FFN hidden dimension:
\begin{equation}
\widetilde W^{\mathrm{gate}}_{l,e}=P_{l,e}W^{\mathrm{gate}}_{l,e},
\quad
\widetilde W^{\mathrm{up}}_{l,e}=P_{l,e}W^{\mathrm{up}}_{l,e},
\quad
\widetilde W^{\mathrm{down}}_{l,e}
=
W^{\mathrm{down}}_{l,e}P_{l,e}^{\top}.
\label{eq:reorder_weights}
\end{equation}

After reordering, a retention ratio $r\in(0,1]$ corresponds to keeping only the top $k=\lceil r\, d_{\mathrm{ff}}\rceil$ channels in all ranked channels.
Let $m(r)$ denote the corresponding prefix mask. The sliced expert FFN is
\begin{equation}
\mathrm{FFN}^{(r)}_{l,e}(h)
=
\widetilde W^{\mathrm{down}}_{l,e}
\Big(
m(r)\odot
\phi(\widetilde W^{\mathrm{gate}}_{l,e}h)
\odot
\widetilde W^{\mathrm{up}}_{l,e}h
\Big).
\label{eq:prefix_slice}
\end{equation}
During slice action learning, masks are only used to simulate channel removal and mark action training results (which channels are retained), while all ranked model weights remain unchanged; at deployment, masked channels can instead be truly removed or excluded from computation.
The resulting importance-ordered layout forms a shared prefix-sliceable weight space from which different expert widths can be selected across pruning budgets.

\subsection{Action Mask Learning}

Given the ranked experts, we learn one discrete retention action per expert from predefined ratio set 
$\mathcal A=\{r_1,\ldots,r_K\}$, where $r_k\in(0,1]$.
For every expert $(l,e)$, we maintain trainable action logits $\alpha_{l,e}\in\mathbb R^K$. To sample subnetworks under one-hot operation while preserving gradients for action logits training, we use Straight-Through Gumbel-Softmax \citep{jang2017gumbel,fang2024maskllm}:
\begin{equation}
z^{\mathrm{soft}}_{l,e}
=
\mathrm{softmax}\!\left(\frac{\alpha_{l,e}+g_{l,e}}{\tau}\right),
\qquad
z^{\mathrm{hard}}_{l,e}
=
\mathrm{one\_hot}\!\left(\arg\max_k z^{\mathrm{soft}}_{l,e,k}\right).
\label{eq:gumbel_action}
\end{equation}

where $g_{l,e,k}\sim \mathrm{Gumbel}(0,1)$ are added to the original action distribution as action noise sampling, and introduce a linearly annealed $\tau$ to control the degree of impact of Gumbel noise to promote early stage action exploration. The sampled expert subnet forward pass uses the straight-through estimator
\begin{equation}
\tilde z_{l,e}
=
\mathrm{sg}\!\left[z^{\mathrm{hard}}_{l,e}-z^{\mathrm{soft}}_{l,e}\right]
+
z^{\mathrm{soft}}_{l,e},
\label{eq:st_action}
\end{equation}
where $\mathrm{sg}(.)$ means stop gradient. $\tilde z_{l,e}$ ensures actions are discrete in the forward pass but differentiable in the backward pass. All straight-through estimator  $\tilde z$  jointly defines a sampled action mask during training stage which can be applied to each expert and produce a sampled subnet nested in full ranked MoE model.

Importantly, we freeze all pretrained model weights during action learning and optimize only the action logits. 
This is essential to our one-for-all design: if model weights were updated together with the pruning actions, masks exported at different points of the training trajectory would correspond to different underlying weight states, breaking the shared-weight property across budgets. 
By keeping the ranked weights fixed, all exported action masks are optimized and will operate on the same parameter set, allowing a single action-training run to produce a consistent multi-budget subnet family nested in full ranked MoE model.

We optimize the action logits using
\begin{equation}
\mathcal L_{\mathrm{action}}
=
\mathcal L_{\mathrm{qual}}
+
\lambda_{\mathrm{cost}}(t)
\sum_{l=1}^{L}\sum_{e=1}^{E}
q_{l,e}
\sum_{k=1}^{K}
p_{l,e,k}r_k
-
\beta(t)\mathcal H(p),
\label{eq:action_objective}
\end{equation}

$\mathcal L_{\mathrm{qual}}$ combines the student language-modeling cross-entropy with teacher--student KL distillation from the frozen full model. $p_{l,e}=\mathrm{softmax}(\alpha_{l,e})$ denotes the currently learned clean action distribution without Gumbel sampling noise and $\tau$ scaling. $q_{l,e}$ is the frequency of assigning tokens to expert $(l,e)$ in this MoE layer. This cost term is both parameter and load weighted: assigning thicker actions to highly routed experts preserves task accuracy, but also brings more computation and incurs larger cost penalties, so actions must learn an expert-wise accuracy--efficiency trade-off. We further include an entropy regularizer $\mathcal H(p)$ computed from all $p_{l,e}$. It's the mean action categorical entropy over all experts averaged across layers and experts. This term encourages exploration, prevents premature collapse to a single action, and helps the model discover more reliable MoE subnets. Entropy weight $\beta(t)$ is also annealed linearly. 

Action logits are initialized as the fully retained action.
We gradually increase $\lambda_{\mathrm{cost}}(t)$ so that the learned policy moves from thicker to thinner experts and traverses progressively stronger pruning budgets.
At any checkpoint, the training have:
\begin{equation}
\hat r_{l,e}
=
r_{\arg\max_k\alpha_{l,e,k}},
\qquad
\hat{\mathcal M}
=
\{\hat r_{l,e}\}_{l,e},
\qquad
\hat\rho
=
1-
\frac{1}{LE}
\sum_{l=1}^{L}\sum_{e=1}^{E}
\hat r_{l,e}.
\label{eq:export_mask_and_budget}
\end{equation}
Here, $\hat r_{l,e}$ is the currently trained deterministic expert action, $\hat{\mathcal M}$ specifies one retained prefix mask for each expert and $\hat\rho$ is the global fraction of routed-expert parameters pruned.
By saving masks $\hat{\mathcal M}$ along this optimization trajectory, FlexMoE obtains a sequence of pruned subnetworks across budgets in one training run, and marked by trained masks without actually store separated model checkpoints.


\textbf{FlexMoE's Kernel Co-Design.}
Shared-weight online budget switching adds two runtime costs absent from
statically pruned subnetworks: the expert-specific retained widths make the
expert FFNs irregular, fragmented GEMMs, and reconstructing each clipped weight
by slicing and concatenating the gate--up tensor adds per-call tensor
manipulation and host-side dispatch. During decoding these are dominated by
kernel-launch and scheduling overhead that does not shrink with the
reduced expert width, so the theoretical FLOP savings of pruning stay hidden. We
therefore co-design a clipped-FFN kernel that (i) stores the gate--up weights
interleaved and the down projection transposed, so a retained prefix weight is selected
by loop bounds and masking rather than any weight slicing, copy or concatenation, and
(ii) turns the routing into fixed-shape, device-resident block metadata and
reads each expert's retained width from a device buffer inside the kernel,
pruning on-device---skipping pruned channel blocks and clamping the contraction
loop---while the launch shape stays constant. Eliminating every host
synchronization and data-dependent branch makes the whole expert forward
capturable in a single CUDA graph; decoding then becomes compute-bound, so the
smaller active width turns into real speedup, and switching the budget only
overwrites the tiny device width buffer, requiring no weight reload and no graph
re-capture. See Appendix~B for implementation details.

\subsection{Recovery Fine-Tuning}

Parameter pruning can still degrade model quality, while independently recovering every budget would undermine the one-for-all objective.
We therefore choose one mid-budget mask $\mathcal M^{\mathrm{mid}}$ and perform a single-point recovery fine-tune.
The importance-ranked base weights $W_0$ are frozen, and LoRA adapters~\citep{hu2022lora} with parameters $\varphi$ are trained on the retained expert FFN channels chosen by $\mathcal M^{\mathrm{mid}}$ at this budget.

Let $o^{S}_{\mathrm{mid}}=f_{\mathrm{stu}}(x;\mathcal M^{\mathrm{mid}},W_0,\varphi)$ denote the outputs from LoRA-augmented subnet materialized by mask $\mathcal M^{\mathrm{mid}}$
and $o^{F}=f_{\mathrm{tea}}(x;W_0)$ denote the full-model teacher outputs, respectively.
We optimize
\begin{equation}
\mathcal L_{\mathrm{rec}}
=
\mathbb E_{(x,y)\sim\mathcal D_{\mathrm{rec}}}
\left[
\lambda_{\mathrm{task}}\mathrm{CE}(o^{S}_{\mathrm{mid}},y)
+
\lambda_{\mathrm{kl}}\mathrm{KL}(o^{F}\Vert o^{S}_{\mathrm{mid}})
\right].
\label{eq:recovery_objective}
\end{equation}
to let the pruned sub-network mimic the full teacher model. After training, the adapters are merged into the ranked base model to obtain one recovered model $\overline W$.
Because all exported masks select prefixes from the same importance-ordered expert weight space while preserving the router and expert topology, the same $\overline W$ can be reused with every budget mask $\mathcal M^{(b)}$ without additional recovery fine-tuning or budget-specific checkpoints.
Thus, the one-for-all property extends from action learning to recovery.

\section{Experiments}

\subsection{Experimental Setup}
\label{sec:exp-setup}
\paragraph{Implementation and Evaluation Details.}

We use three pretrained MoE LLMs: Mixtral-8x7B~\citep{jiang2024mixtral}, Phi-3.5-MoE~\citep{abdin2024phi3} and Qwen2-57B-A14B~\citep{yang2024qwen2} as cross-model validation with different MoE architectures and sizes. We defined discrete action set as $\mathcal A=\{0.1, 0.4,0.7,1\}$, and throughout this section, the pruning ratio (budget) is defined by the global prune budget $\hat{\rho}$ in Eq.~\eqref{eq:export_mask_and_budget} of an action mask applied to the ranked full MoE model. During action learning, we export one action mask whenever expert parameters globally pruned by roughly $1\%$ ($\hat{\rho}$ increase $1\%$). For recovery fine-tuning, we choose the $40\%$ budget action mask as the recovery point, and let recovered ability transfer to other budgets (20\% and 60\% here). Channel importance ranking, action learning, and recovery fine-tuning stages use Zyda-2~\citep{tokpanov2024zyda2} as the calibration dataset. As quality evaluation, we report zero-shot accuracy on seven widely used reasoning benchmarks implemented with \texttt{lm-eval-harness}: ARC-Challenge, ARC-Easy, HellaSwag, OpenBookQA, PIQA, WinoGrande, and MathQA~\citep{gao2021framework,clark2018arc,zellers2019hellaswag,mihaylov2018openbookqa,bisk2020piqa,sakaguchi2020winogrande,amini2019mathqa}. For each model, we independently rerun FlexMoE pipeline three times using different random seeds to account for randomness in calibration-sample selection and parameter initialization, and report the mean and sample standard deviation. More hyper-parameters and reproduction details please see Appendix~\ref{app:hyperparameters}.


\paragraph{Baselines.}
We fix the baseline family throughout the main comparison. Our primary baseline methods are \moesvd{} and TD-MoE, since both are latest strong pretrained MoE compression methods \citep{li2025moesvd,xu2026tdmoe}. We additionally report \naee{} and \moeiitwo{} as broader references, representing expert-level pruning/skipping and mixed inter-/intra-expert compression, respectively \citep{lu2024naee,yang2024moei2}. For fair comparison, we locally reproduced the two strongest baseline methods, MoE-SVD and TD-MoE on all 3 backbones by using same Zyda-2 calibration dataset, same local \texttt{lm-eval-harness} protocal over 7 tasks. To avoid selective reporting, we use a shared comparison grid centered on 20\%, 40\%, and 60\% prune budgets, and include each baseline whether the corresponding model--budget pair is publicly available or we locally reproduced. When a baseline is still unavailable for a given model--budget pair, we marked as "N/A".

\subsection{Main Results and Analysis}

\begin{table*}[t]
\centering
\label{tab:main_results}

\scriptsize
\setlength{\tabcolsep}{2pt}
\renewcommand{\arraystretch}{1.3}

\resizebox{\textwidth}{!}{%
\begin{tabular}{c l c c c c c c c c c c}
\toprule
\multirow{2}{*}{\textbf{Prune}} &
\multirow{2}{*}{\textbf{Method}} &
\multicolumn{8}{c}{\textbf{Mixtral-8$\times$7B}} &
\multicolumn{1}{c}{\textbf{Phi-3.5-MoE}} &
\multicolumn{1}{c}{\textbf{Qwen2-57B-A14B}} \\
\cmidrule(lr){3-10}
\cmidrule(lr){11-11}
\cmidrule(lr){12-12}
& &
\textbf{ARC-c} &
\textbf{ARC-e} &
\textbf{HellaS} &
\textbf{OBQA} &
\textbf{PIQA} &
\textbf{WinoG} &
\textbf{MathQA} &
\textbf{Avg. (SD)} &
\textbf{Avg. (SD)} &
\textbf{Avg. (SD)} \\
\midrule

0\% & Base model
& 57 & 84 & 65 & 36 & 82 & 76 & 43
& 63.29 & 62.00 & 58.71 \\

\midrule

\multirow{6}{*}{20\%}

& NAEE\citeyear{lu2024naee}
& 47 & 76 & 58 & 32 & 79 & 72 & 40
& 57.71 & N/A & 55.86 \\

& MoE-I$^2$\citeyear{yang2024moei2}
& 48 & 79 & 55 & 32 & 78 & 74 & 37
& 57.57 & N/A & N/A \\

& MoE-SVD\textsuperscript{*}\citeyear{li2025moesvd}
& 50 & 80 & 60 & 32 & 80 & 71 & 35
& 58.29 & 60.57 & 58.43 \\

& TD-MoE\textsuperscript{*}\citeyear{xu2026tdmoe}
& 47 & 77 & 56 & 28 & 79 & 74 & 36
& 56.71 & 58.43 & 58.00 \\

\midrule

& FlexMoE
& \stat{52.00}{2.00}
& \stat{80.00}{2.00}
& \stat{63.67}{0.58}
& \stat{35.67}{0.58}
& \stat{80.00}{1.73}
& \stat{77.00}{1.00}
& \stat{40.00}{1.00}
& \stat{61.19}{0.92}
& \stat{60.24}{0.36}
& \stat{58.81}{0.16} \\

& FlexMoE(SP-XFER)
& \stat{53.67}{1.53}
& \stat{81.33}{1.15}
& \stat{64.33}{0.58}
& \stat{35.67}{0.58}
& \stat{80.33}{1.15}
& \stat{77.00}{1.00}
& \stat{40.00}{1.00}
& \stat{\textbf{61.76}}{0.86}
& \stat{\textbf{63.29}}{0.00}
& \stat{\textbf{60.24}}{0.7} \\

\midrule

\multirow{6}{*}{40\%}

& NAEE\citeyear{lu2024naee}
& 36 & 63 & 46 & 25 & 72 & 64 & 35
& 48.71 & 57.57 & 53.14 \\

& MoE-I$^2$\citeyear{yang2024moei2}
& 38 & 71 & 43 & 26 & 69 & 66 & 31
& 49.14 & 45.29 & N/A \\

& MoE-SVD\textsuperscript{*}\citeyear{li2025moesvd}
& 38 & 72 & 49 & 27 & 75 & 65 & 27
& 50.43 & 55.57 & 54.86 \\

& TD-MoE\textsuperscript{*}\citeyear{xu2026tdmoe}
& 30 & 57 & 39 & 20 & 66 & 62 & 24
& 42.57 & 55.14 & 55.43 \\

\midrule

& FlexMoE
& \stat{45.00}{1.00}
& \stat{75.67}{0.58}
& \stat{55.33}{1.53}
& \stat{31.67}{1.53}
& \stat{78.33}{0.58}
& \stat{71.33}{0.58}
& \stat{31.00}{1.00}
& \stat{55.48}{0.87}
& \stat{55.43}{0.43}
& \stat{58.86}{0.14} \\

& FlexMoE(SP-FT)
& \stat{48.33}{1.15}
& \stat{77.00}{1.00}
& \stat{60.00}{1.00}
& \stat{33.67}{1.15}
& \stat{79.67}{0.58}
& \stat{73.33}{0.58}
& \stat{33.67}{0.58}
& \stat{\textbf{57.95}}{0.5}
& \stat{\textbf{59.62}}{0.33}
& \stat{\textbf{60.00}}{0.42} \\

\midrule

\multirow{6}{*}{60\%}

& NAEE\citeyear{lu2024naee}
& 23 & 42 & 33 & 17 & 62 & 55 & 26
& 36.86 & N/A & 44.00 \\

& MoE-I$^2$\citeyear{yang2024moei2}
& 22 & 44 & 32 & 18 & 58 & 55 & 23
& 36.00 & N/A & N/A \\

& MoE-SVD\textsuperscript{*}\citeyear{li2025moesvd}
& 28 & 60 & 40 & 22 & 66 & 62 & 25
& 43.29 & 50.00 & 51.71 \\

& TD-MoE\textsuperscript{*}\citeyear{xu2026tdmoe}
& 22 & 37 & 29 & 16 & 58 & 54 & 22
& 34.00 & 46.14 & 51.43 \\

\midrule

& FlexMoE
& \stat{31.33}{1.53}
& \stat{61.33}{2.31}
& \stat{46.33}{0.58}
& \stat{23.00}{1.00}
& \stat{73.33}{0.58}
& \stat{65.00}{2.65}
& \stat{26.00}{1.00}
& \stat{46.62}{0.81}
& \stat{50.43}{0.25}
& \stat{58.62}{0.08} \\

& FlexMoE(SP-XFER)
& \stat{34.33}{2.08}
& \stat{64.33}{3.05}
& \stat{50.33}{1.15}
& \stat{23.33}{0.58}
& \stat{73.67}{1.53}
& \stat{70.33}{0.58}
& \stat{26.33}{1.53}
& \textbf{\stat{\textbf{48.95}}{0.84}}
& \textbf{\stat{\textbf{52.81}}{0.22}}
& \textbf{\stat{\textbf{59.76}}{0.58}} \\

\bottomrule
\end{tabular}%
}
\caption{
Zero-shot accuracy (\%) across seven downstream tasks.
FlexMoE results report mean $\pm$ sample standard deviation over three
independent full-pipeline runs.
$^{*}$ denotes locally reproduced method results.
We apply single-point recovery fine-tuning (SP-FT) at the 40\% budget;
the 20\% and 60\% results reuse the same recovered weights (SP-XFER)
without additional fine-tuning.
}
\end{table*}

\paragraph{Task Accuracy Results Analysis.}


Across models, the main practical pattern is consistent: FlexMoE achieves competitive or superior pruning quality without recovery across most evaluated settings. With a single shared recovery fine-tuning run, it consistently surpasses all strong baselines, and the repeated runs support that FlexMoE preserves quality with varied initialization and sample batches. For recovery strategy, the shared weight fine-tuned at the 40\% point transferred recovery to other (20\% and 60\%) budget points and achieves higher task accuracy.  Observing the excellent performance of Qwen2, in addition to comparing it with the baseline, we also tested the performance on Qwen2 at compression ratios of 50\% and 80\% (results see appendix Table~\ref{tab:full-results}). Surprisingly, without fine-tuning recovery, average scores show that it remains nearly lossless at a $50\%$ parameter prune budget ($\sim 99.8\%$ of base performance), even at an $80\%$ prune budget, it still retains about $92.9\%$ of the base average score. These results demonstrate the effectiveness of FlexMoE in pruning quality preservation. These results confirming that our proposed expert-wise action learning captures meaningful architecture-dependent structure in MoE layers rather than random slicing allocation and budget compression.

\begin{figure}
    \centering
    \includegraphics[width=0.9\textwidth]{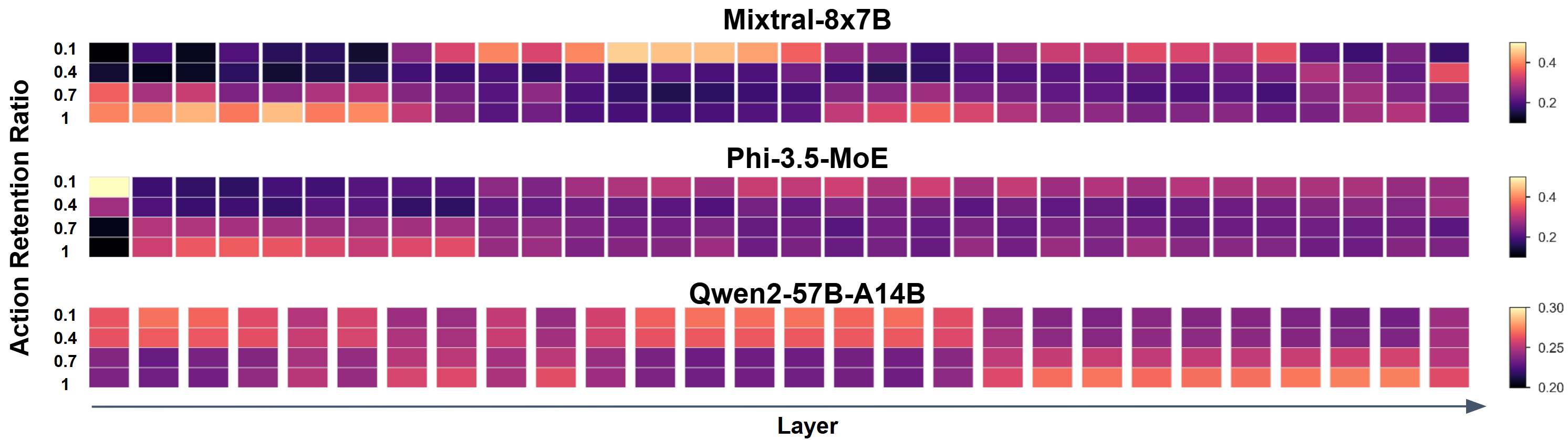}
    \caption{Layer-wise action distributions learned by FlexMoE at 40\% prune budget. Lighter cell color indicate the higher fraction of experts in that layer assigned to the corresponding action.}
    \label{fig:cross_model_action_profile}
\end{figure}

\paragraph{Effectiveness of Action Learning.}
Figure~\ref{fig:cross_model_action_profile} and Figure~\ref{fig:ablation_action_mask_example} jointly show that the learned action distributions are clearly non-uniform across experts, layers and models, indicating that FlexMoE does not merely prune experts, but learns structured budget allocation. Table~\ref{tab:ablation_action_shuffle} shows results for further ablation experiments of slicing action assignment. The learned action assignment achieves the highest mean accuracy across 4 model-budget settings, outperforming all uniform, Gaussian-random, globally shuffled, or in-layer-shuffled action assignment control. Since global shuffling preserves the overall prune ratio and in-layer shuffling additionally preserves layer-wise capacity, their degradation indicating that destroying learned action structure consistently hurts performance, and expert-specific assignments matter. These results strongly confirm that our proposed expert-wise action learning captures meaningful quality-dependent structure rather than random budget convergence and allocation, which is the key to FlexMoE maintaining strong quality across pruning budgets.

\paragraph{Effectiveness of Single-Point Fine-Tuning.}
Table~\ref{tab:spft_cross_budget_transfer} shows that the quality gains
obtained by SP-FT transfer effectively to both tighter and looser pruning
budgets. This transfer follows from the shared
importance-ordered prefix structure: tighter budget action masks retain smaller subsets
of the channels updated in the recovery fine-tuning, whereas looser masks
preserve these recovered prefixes while additionally exposing more tail channels not attended in recovery.

\begin{wraptable}[12]{r}{0.54\linewidth}
\vspace{-12.5pt}
\centering
\fontsize{8pt}{10pt}\selectfont
\renewcommand{\arraystretch}{0.9}
\caption{
Cross-budget transfer ability of SP-FT recovered weights on
Mixtral-8$\times$7B. Values are seven-task average accuracies (\%).
The last column independently fine-tunes at each test budget ratio.
}
\resizebox{\linewidth}{!}{%
\begin{tabular}{l c c c c}
\toprule
\multirow{2}{*}{\vspace{10pt}\textbf{Test Ratio}} &
\shortstack{\textbf{SP-FT}\\\textbf{at 20\%}} &
\shortstack{\textbf{SP-FT}\\\textbf{at 40\%}} &
\shortstack{\textbf{SP-FT}\\\textbf{at 60\%}} &
\shortstack{\textbf{Independent}\\\textbf{per-budget FT}} \\
\midrule

20\%
& 60.86
& \textbf{61.86}
& 60.71
& 60.86 \\

40\%
& 54.57
& \textbf{58.00}
& 57.14
& \textbf{58.00} \\

60\%
& 42.86
& 49.57
& \textbf{50.29}
& \textbf{50.29} \\

\midrule

Average
& 52.76
& \textbf{56.48}
& 56.05
& 56.38 \\

\bottomrule
\end{tabular}%
}

\label{tab:spft_cross_budget_transfer}
\end{wraptable}
We also present an ablation of SP-FT at different budget points. Among three recovery points,
SP-FT at 40\% achieves the highest cross-budget average score
compared with recovery at 20\% and 60\%. More broadly, the shared model recovered at 40\% slightly exceeds the average performance of independently fine-tuning each budget, while requiring only one recovery run and one recovered weight set. Thus, a single point recovery fine-tune produces
one shared weight that remains strong quality across the budget trajectory,
avoiding repeated per-budget recover and the storage of separate
budget-specific weights. Together with one-run slice action learning, this
directly supports FlexMoE's train-once, deploy-many objective.

\begin{figure}[t]
\centering
\begin{minipage}[t]{0.4\linewidth}
\centering
\captionof{figure}{
Learned action mask for Mixtral-8$\times$7B at 40\% pruning.
Each cell denotes one expert's selected retention ratio.
}
\label{fig:ablation_action_mask_example}
\includegraphics[
    width=\linewidth
]{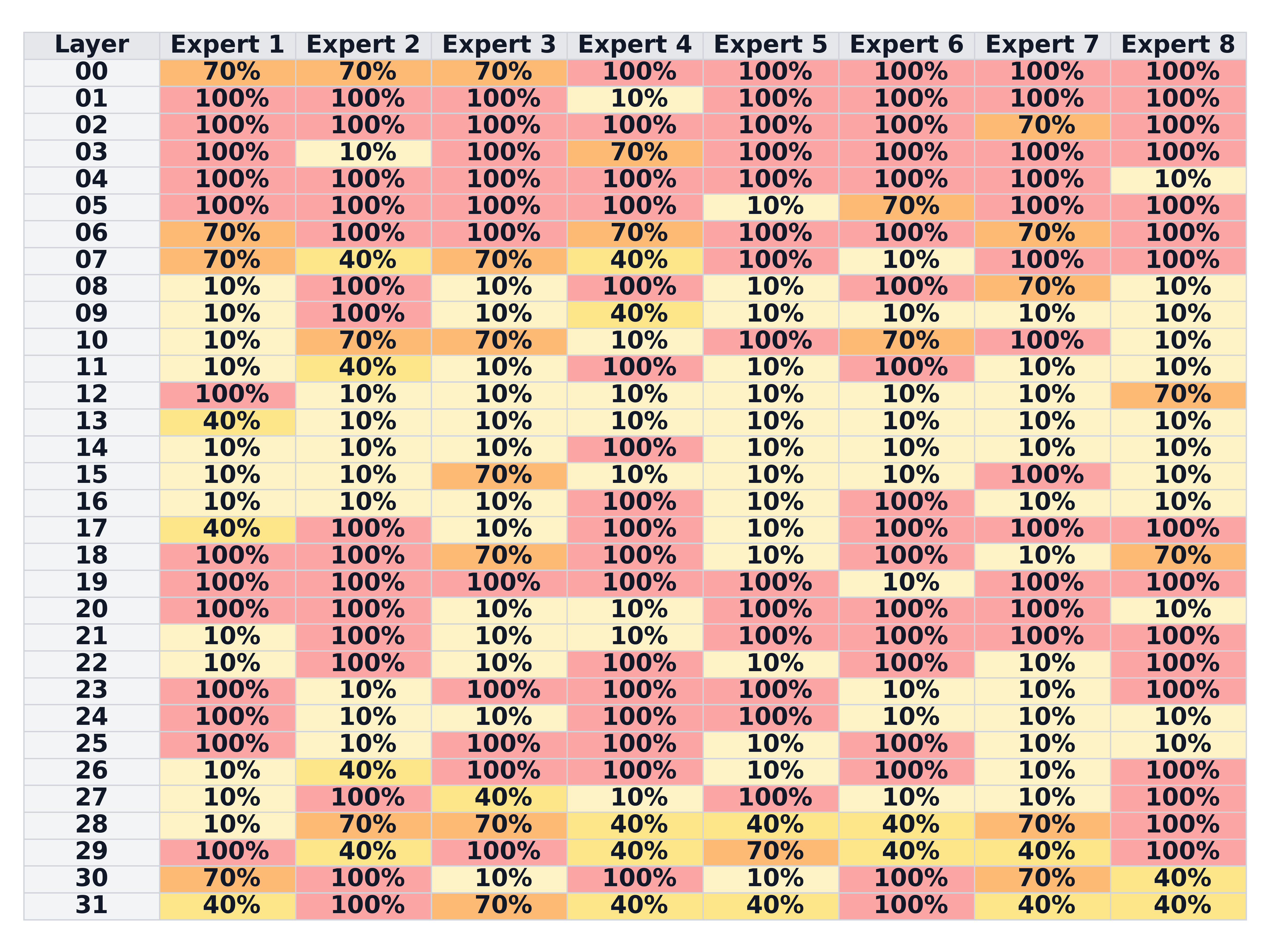}

\captionsetup{
    type=figure,
    font=small,
    skip=3pt
}
\end{minipage}
\hspace{0.015\linewidth}
\begin{minipage}[t]{0.53\linewidth}

\setlength{\tabcolsep}{1pt}
\renewcommand{\arraystretch}{1.5}
\captionof{table}{
Expert-wise action-allocation ablation.
Values are seven-task average zero-shot accuracies (\%).
$\pm$ denotes sample SD; 
}\vspace{15pt}
\fontsize{8pt}{8.5pt}\selectfont
\label{tab:ablation_action_shuffle}
\begin{tabular}{l c c c c}
\toprule
\multirow{2}{*}{\textbf{Allocation}} &
\multicolumn{2}{c}{\textbf{Mixtral-8$\times$7B}} &
\multicolumn{2}{c}{\textbf{Qwen2-57B-A14B}} \\
\cmidrule(lr){2-3}
\cmidrule(lr){4-5}
& \textbf{20\%} & \textbf{60\%}
& \textbf{20\%} & \textbf{60\%} \\
\midrule

Gaussian random
& \stat{60.00}{0.68}
& \stat{44.14}{2.08}
& \stat{57.86}{0.94}
& \stat{57.00}{1.21} \\

Uniform assign
& 60.14
& 45.14
& 58.14
& 57.43 \\

Learned shuffled
& \stat{59.14}{0.55}
& \stat{44.14}{1.84}
& \stat{55.71}{0.46}
& \stat{56.14}{0.57} \\

In-layer shuffled
& \stat{60.43}{0.44}
& \stat{46.00}{0.21}
& \stat{58.29}{0.13}
& \stat{58.43}{0.0} \\

Learned
& \textbf{\stat{61.19}{0.92}}
& \textbf{\stat{46.62}{0.81}}
& \textbf{\stat{58.81}{0.16}}
& \textbf{\stat{58.62}{0.08}} \\

\bottomrule
\end{tabular}

\captionsetup{
    type=table,
    font=small,
    skip=3pt
}
\vspace{13pt}

\end{minipage}
\end{figure}

%

\subsection{Deployment Performance and Analysis}
\label{sec:deploy}

We finally evaluated deployment performance of FlexMoE's pruned models under 2 scenarios: Fixed hardware budget and under MoE kernel co-design. We evaluate these two settings separately to distinguish their sources of throughput
improvement. We use the SGLang engine~\citep{zheng2024sglang} on a single H200 GPU under a synthetic workload with 4096 prompt requests, input length $=64$, output length $=256$.
Our primary metric is model throughput (tok/s), defined by summarizing the model prefill input and decoding output throughput. Results are averaged across multiple runs $\pm$ sample standard
deviation, and speedups are normalized to the unpruned model within each execution path.

\paragraph{Fixed hardware budget.}
For deployments with a fixed GPU-memory budget, we materialize each target
subnetwork by physically removing the pruned expert channels at that budget. This provides
three coupled benefits: it reduces the resident weight footprint, decreases
the FLOPs and weight traffic of active experts, and allows the serving engine
to allocate the released memory to additional KV-cache slots. The resulting
throughput therefore represents the practical end-to-end benefit available
under the same deployment (GPU-memory) budget. As shown in Table~\ref{tab:static_deployment_throughput}, static
pruned model consistently improves throughput across all three
architectures. 

This fixed-memory-budget experiment reflects realistic deployment: 
parameter memory released through compression is naturally available to the serving 
engine for KV-cache allocation. Thus, the KV-cache benefit is not an unrelated artifact, 
but an inherent deployment advantage of reducing the checkpoint footprint.



\begin{table*}[t]

\centering
\caption{
End-to-end serving throughput and memory allocation of statically
materialized FlexMoE subnetworks under a fixed GPU-memory budget.
}
\label{tab:static_deployment_throughput}
\fontsize{8pt}{10pt}\selectfont
\setlength{\tabcolsep}{0.3pt}

\begin{tabular*}{\textwidth}{
@{\extracolsep{\fill}}
c
ccc
ccc
ccc
@{}
}
\toprule

 &
\multicolumn{3}{c}{\textbf{Mixtral-8$\times$7B}} &
\multicolumn{3}{c}{\textbf{Phi-3.5-MoE}} &
\multicolumn{3}{c}{\textbf{Qwen2-57B-A14B}} \\

\cmidrule(lr){2-4}
\cmidrule(lr){5-7}
\cmidrule(lr){8-10}

Pruning Ratio &
tok/s &
ckpt size (GB) &
KV (GB) &

tok/s &
ckpt size  &
KV  &

tok/s &
ckpt size  &
KV  \\

\midrule

0\% &
\stat{8276}{67} & 87.03 & 24.26 &
\stat{8226}{112} & 78.05 & 33.24 &
\stat{1302}{9} & 107.02 & 4.27 \\

20\% &
\stat{10478}{115} & 70.13 & 41.15 &
\stat{9761}{128} & 63.08 & 48.21 &
\stat{4455}{18} & 88.62 & 22.67 \\

40\% &
\stat{11972}{107} & 53.56 & 57.73 &
\stat{11341}{134} & 48.05 & 63.24 &
\textbf{\stat{5214}{32}} & 70.21 & 41.08 \\

60\% &
\textbf{\stat{12296}{135}} & 37.09 & 74.20 &
\textbf{\stat{11802}{148}} & 36.61 & 74.68 &
\stat{5185}{44} & 51.95 & 59.34 \\

\bottomrule
\end{tabular*}
\end{table*}

\begin{table*}[t]
\centering
\caption{
Inference throughput with FlexMoE co-designed kernel enabled, compared with naive Python inference implementation on Mixtral.
}
\label{tab:online_kernel_throughput}

\fontsize{8pt}{10pt}\selectfont
\setlength{\tabcolsep}{1pt}

\begin{tabular*}{\textwidth}{
@{\extracolsep{\fill}}
c ccc cc cc
@{}
}
\toprule

Pruning &
\multicolumn{3}{c}{\textbf{Mixtral-8$\times$7B}} &
\multicolumn{2}{c}{\textbf{Phi-3.5-MoE}} &
\multicolumn{2}{c}{\textbf{Qwen2-57B-A14B}} \\

\cmidrule(lr){2-4}
\cmidrule(lr){5-6}
\cmidrule(lr){7-8}

Ratio &
\textbf{Naive Python tok/s} &
\textbf{Kernel tok/s} &
\textbf{speedup$\times$} &
\textbf{Kernel} &
\textbf{speedup} &
\textbf{Kernel} &
\textbf{speedup} \\

\midrule

0\% &
7780&
\stat{5818}{31} & $1.00\times$ &
\stat{10639}{100} & $1.00\times$ &
\stat{7367}{30} & $1.00\times$ \\

20\% &
7345 &
\stat{6860}{52} & $1.18\times$ &
\stat{12395}{110} & $1.17\times$ &
\stat{9093}{28} & $1.23\times$ \\

40\% &
7087 &
\stat{8222}{43} & $1.41\times$ &
\stat{15286}{135} & $1.44\times$ &
\stat{10744}{36} & $1.46\times$ \\

60\% &
6649 &
\textbf{\stat{11895}{73}} & $\mathbf{2.04\times}$ &
\textbf{\stat{18319}{166}} & $\mathbf{1.72\times}$ &
\textbf{\stat{13018}{45}} & $\mathbf{1.77\times}$ \\

\bottomrule
\end{tabular*}
\end{table*}



\begin{table}[t]
\fontsize{9pt}{10pt}\selectfont
\centering
\caption{Cost of switching the serving budget (Qwen2-57B-A14B, H200, bf16).
Our in-place width-buffer swap replaces checkpoint reload and graph re-capture at
$10^4$--$10^6\times$ lower latency, while keeping the shared weights resident and
outputs bit-identical. Reload is a lower bound (weight transfer and reload only under our cluster at a cross-node sequential checkpoint read bandwidth of $\sim10.4$\,GB/s and a host-to-GPU transfer bandwidth of $\sim53.2$\,GB/s.}
\label{tab:switch}
\vspace{7pt}
\begin{tabular}{lccc}
\toprule
budget switch mechanism & Switch latency & Reloads weights & Re-captures graph \\
\midrule
Reload materialized per-budget checkpoint & $6.4$--$13.2$\,s & yes & yes \\
Re-capture CUDA graph per budget          & $108$\,ms        & no  & yes \\
\textbf{In-place width-buffer swap (ours)}& $\mathbf{4.2\,\mu s}$ & no & no \\
\bottomrule
\end{tabular}
\end{table}


\paragraph{Online budget switching with the co-designed kernel.}
We next isolate the computational benefit of activating thinner experts: the full
importance-ordered weights stay resident and the per-expert retained widths are
selected at runtime, so the resident model size, KV-cache capacity, and memory
headroom are identical across budgets. A direct Python implementation cannot
realize the FLOP savings---during decoding the step is dominated by kernel launch
and host-side scheduling that does not shrink with the retained width
(Table~\ref{tab:online_kernel_throughput}, \emph{naive}). Our co-designed kernel removes every host
synchronization and data-dependent host branch, so the entire expert forward is
captured in a single CUDA graph and replayed; decoding then becomes compute-bound
and the reduced active width turns into real speedup. As shown in
Table~\ref{tab:online_kernel_throughput}, the kernel yields consistent throughput gains on all three backbones as expert parameter reduce.

\paragraph{Runtime budget switching is instant and almost free.}
Because the kernel reads each expert's retained width from a device buffer and its
launch shape is width-independent, a single captured graph serves every
budget: switching the budget only overwrites that $O(E\!\cdot\!L)$ width buffer
($\sim7$\,KB). 
\begin{wrapfigure}{r}{0.6\textwidth}
  \vspace{-9pt}
  \centering
  \includegraphics[width=0.6\textwidth]{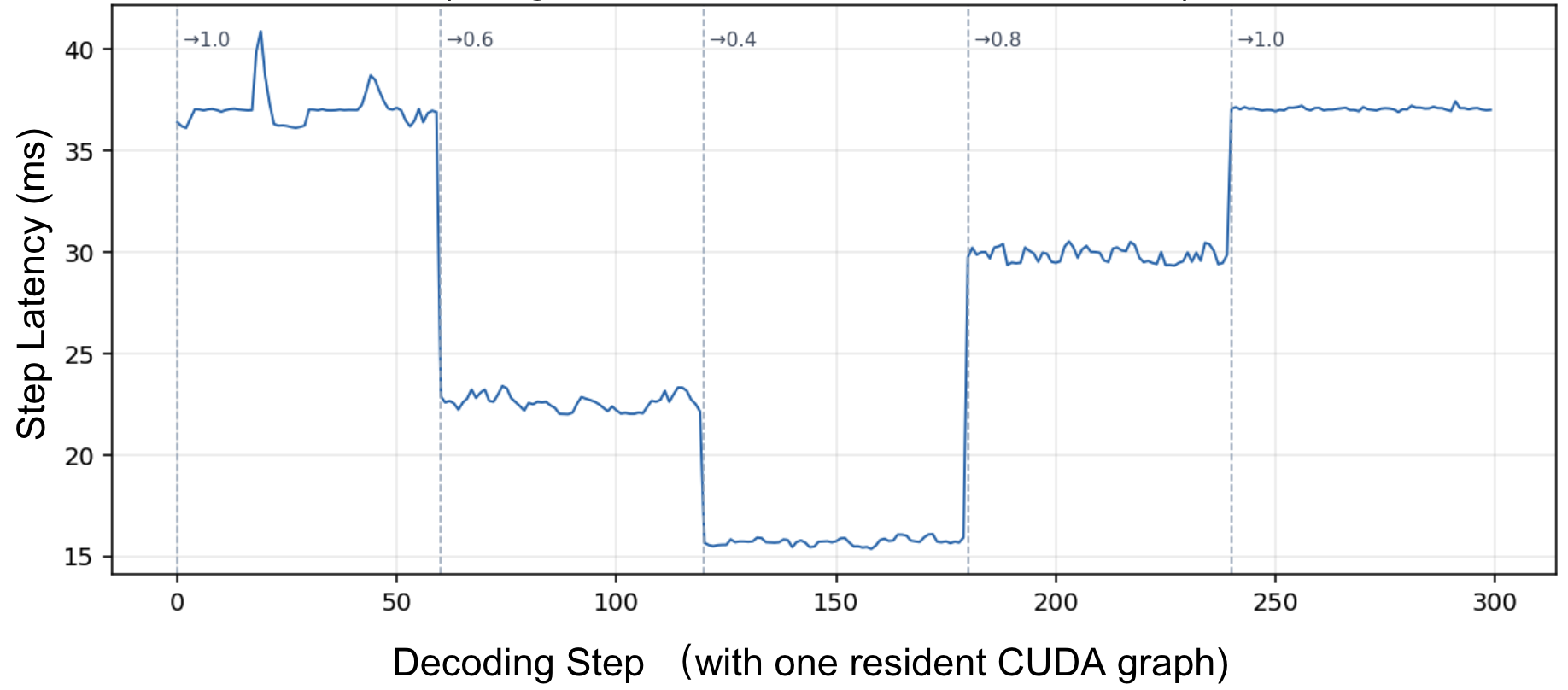}
  \caption{MoE decode-step latency under co-designed kernel with the serving budget scheduled as $1.0\!\to\!0.6\!\to\!0.4\!\to\!0.8\!\to\!1.0$, 60 steps each budget.}
  \label{fig:timeline}
  \vspace{-1.2\baselineskip}
\end{wrapfigure}
This costs $4.2\,\mu s$---four orders of magnitude below re-capturing
the graph ($108$\,ms) and six below reloading a materialized per-budget checkpoint
($6.4$--$13.2$\,s; Table~\ref{tab:switch}). The switched graph is bit-identical
to a graph captured natively at that budget (max.\ abs.\ difference $0$; and
$\le 8\!\times\!10^{-6}$ against an eager reference) and stays within $0.7\%$ of its
latency, so this flexibility costs neither accuracy nor speed, while the MoE decode
step itself shrinks by up to $2.37\times$ as the budget drops. As shown in Figure~\ref{fig:timeline}, under the kernel, decoding step latency tracks the scheduled serving budget instantly and reversibly at fixed resident memory, with no extra graph re-capture and weight reload cost (Qwen2-57B-A14B, H200). 

\section{Conclusion}

We presented FlexMoE, a post-training compression framework that converts a pretrained MoE LLM into a family of deployable subnetworks nested in the full model. By ranking expert FFN channels by estimated importance and learning per-expert one discrete retention action across budgets, our method obtains a series of reliable pruned subnetworks nested in large pretrained MoE from a single action-training run. We further showed that a one-step recovery fine-tune at a single mid-budget point is already sufficient to produce shared recovered model that transfers well to other unseen budgets. Experiments on various MoE models show that the proposed framework surpasses recent strong MoE compression baselines. Finally, offline pruned subnetworks deliver real throughput gains at their fixed budget, and a kernelized co-design makes runtime budget switching feasible. We hope this work provides a new perspective of MoE model structure search and a practical foundation for budget-adaptive MoE model deployment and inference.

\paragraph{Limitations and Future Works.} 
\label{sec:limit}
While our main focus is MoE structure search and static expert-parameter pruning, the multi-budget shared-weight family produced by FlexMoE opens a promising direction for future work on stronger system-level co-design and input-aware dynamic budget-adaptation strategies for deployment and online serving.

\begin{ack}
We would like to acknowledge that computational work involved in this research work is supported
by NUS IT’s Research Computing group using grant numbers CFP03-CF-088. Yang You's research group is being sponsored by ByteDance grant, Alibaba grant, Meta Research Award, AliCloud grant, Apple grant, Google Research Award, Microsoft Research grant, and Adobe gift.
\end{ack}

\bibliographystyle{plainnat}
\bibliography{references}

\clearpage

\appendix

\section{Extended Experimental Results, Ablations and Discussion}
\label{app:extended-results}

\subsection{Full Cross-Model Results, Baseline Comparability, and Multi-Budget Trajectories}
\label{app:full-results}

Table~\ref{tab:full-results} reports the complete cross-model full results among 7 used down-stream tasks.
In addition to the main-text model--budget pairs, 20\%, 40\%, and 60\% points reported in the main text,
we also report additional intermediate budgets
available from the same FlexMoE pruning trajectory using seed 42, to show FlexMoE's robustness across the entire budget trajectory. And we retain source-reported baseline results from their original papers for reference, paired with our locally reproduced
MoE-SVD and TD-MoE results.

\begingroup
\fontsize{6.5pt}{8pt}\selectfont
\setlength{\tabcolsep}{1.5pt}

\begin{longtable}{@{}clcccccccc@{}}
\caption{
Full zero-shot accuracy (\%) across seven downstream tasks.
$\dagger$ denotes publicly available baseline results retained for reference,
while $^{*}$ denotes our local reproduction under the unified calibration
and evaluation setting.
Repeated FlexMoE results report mean $\pm$ sample standard deviation over
three independent full-pipeline runs.
SP-FT denotes recovery at the 40\% budget and SP-XFER reuses the same
recovered weights at another budget. CP-FT denotes the cross-point
fine-tuning ablation, detailed in Appendix~\ref{app:cpft-details}.
}
\label{tab:full-results}\\

\toprule
Ratio & Method
& ARC-c & ARC-e & HellaS & OBQA & PIQA & WinoG & MathQA & Avg. \\
\midrule
\endfirsthead

\multicolumn{10}{c}{\tablename~\thetable\ -- continued} \\
\toprule
Ratio & Method
& ARC-c & ARC-e & HellaS & OBQA & PIQA & WinoG & MathQA & Avg. \\
\midrule
\endhead

\midrule
\multicolumn{10}{r}{Continued on next page} \\
\endfoot

\bottomrule
\endlastfoot

\multicolumn{10}{c}{\textbf{Mixtral-8$\times$7B} (8 experts per layer, top-2 activated per token)} \\
\midrule

0\%
& Base model
& 57 & 84 & 65 & 36 & 82 & 76 & 43 & 63.29 \\

\midrule
20\%
& NAEE$^{\dagger}$
& 47 & 76 & 58 & 32 & 79 & 72 & 40 & 57.71 \\
& MoE-I$^2$$^{\dagger}$
& 48 & 79 & 55 & 32 & 78 & 74 & 37 & 57.57 \\
& MoE-SVD$^{\dagger}$
& 55 & 80 & 61 & 33 & 81 & 73 & 38 & 60.14 \\
& MoE-SVD$^{*}$
& 50 & 80 & 60 & 32 & 80 & 71 & 35 & 58.29 \\
& TD-MoE$^{\dagger}$
& 53 & 83 & 64 & 33 & 82 & 77 & 40 & 61.71 \\
& TD-MoE$^{*}$
& 47 & 77 & 56 & 28 & 79 & 74 & 36 & 56.71 \\
\cmidrule(lr){2-10}
& FlexMoE
& \stat{52.00}{2.00}
& \stat{80.00}{2.00}
& \stat{63.67}{0.58}
& \stat{35.67}{0.58}
& \stat{80.00}{1.73}
& \stat{77.00}{1.00}
& \stat{40.00}{1.00}
& \stat{61.19}{0.92} \\
& Ours (CP-FT) & 53 & 81 & 62 & 33 & 81 & 76 & 37 & 60.43 \\
& FlexMoE (SP-XFER)
& \stat{53.67}{1.53}
& \stat{81.33}{1.15}
& \stat{64.33}{0.58}
& \stat{35.67}{0.58}
& \stat{80.33}{1.15}
& \stat{77.00}{1.00}
& \stat{40.00}{1.00}
& \stat{61.76}{0.86} \\

\midrule
25\%
& FlexMoE (CP-FT)
& 48 & 79 & 61 & 33 & 80 & 75 & 36 & 58.86 \\
& FlexMoE (SP-XFER)
& 53 & 81 & 63 & 34 & 80 & 75 & 38 & 60.57 \\

\midrule
40\%
& NAEE$^{\dagger}$
& 36 & 63 & 46 & 25 & 72 & 64 & 35 & 48.71 \\
& MoE-I$^2$$^{\dagger}$
& 38 & 71 & 43 & 26 & 69 & 66 & 31 & 49.14 \\
& MoE-SVD$^{\dagger}$
& 38 & 72 & 43 & 27 & 71 & 67 & 32 & 50.00 \\
& MoE-SVD$^{*}$
& 38 & 72 & 49 & 27 & 75 & 65 & 27 & 50.43 \\
& TD-MoE$^{\dagger}$
& 47 & 77 & 57 & 28 & 79 & 76 & 35 & 57.00 \\
& TD-MoE$^{*}$
& 30 & 57 & 39 & 20 & 66 & 62 & 24 & 42.57 \\
\cmidrule(lr){2-10}
& FlexMoE
& \stat{45.00}{1.00}
& \stat{75.67}{0.58}
& \stat{55.33}{1.53}
& \stat{31.67}{1.53}
& \stat{78.33}{0.58}
& \stat{71.33}{0.58}
& \stat{31.00}{1.00}
& \stat{55.48}{0.87} \\
& Ours (CP-FT) & 44 & 75 & 57 & 31 & 79 & 68 & 31 & 55.00 \\
& FlexMoE (SP-FT)
& \stat{48.33}{1.15}
& \stat{77.00}{1.00}
& \stat{60.00}{1.00}
& \stat{33.67}{1.15}
& \stat{79.67}{0.58}
& \stat{73.33}{0.58}
& \stat{33.67}{0.58}
& \stat{57.95}{0.5} \\

\midrule
45\%
& FlexMoE (CP-FT)
& 43 & 74 & 56 & 30 & 78 & 72 & 28 & 54.43 \\
& FlexMoE (SP-XFER)
& 44 & 76 & 58 & 31 & 78 & 72 & 31 & 55.71 \\

\midrule
50\%
& MoE-SVD$^{\dagger}$
& 37 & 67 & 50 & 25 & 73 & 64 & 28 & 49.14 \\
\cmidrule(lr){2-10}
& FlexMoE (CP-FT)
& 43 & 73 & 56 & 27 & 78 & 67 & 27 & 53.00 \\
& FlexMoE (SP-XFER)
& 39 & 71 & 56 & 30 & 78 & 71 & 29 & 53.43 \\

\midrule
55\%
& FlexMoE (CP-FT)
& 35 & 66 & 48 & 28 & 75 & 63 & 26 & 48.71 \\
& FlexMoE (SP-XFER)
& 36 & 69 & 54 & 28 & 77 & 69 & 28 & 51.57 \\

\midrule
60\%
& NAEE$^{\dagger}$
& 23 & 42 & 33 & 17 & 62 & 55 & 26 & 36.86 \\
& MoE-I$^2$$^{\dagger}$
& 22 & 44 & 32 & 18 & 58 & 55 & 23 & 36.00 \\
& MoE-SVD$^{\dagger}$
& 23 & 45 & 33 & 19 & 62 & 55 & 25 & 37.43 \\
& MoE-SVD$^{*}$
& 28 & 60 & 40 & 22 & 66 & 62 & 25 & 43.29 \\
& TD-MoE$^{\dagger}$
& 28 & 55 & 38 & 21 & 65 & 62 & 24 & 41.86 \\
& TD-MoE$^{*}$
& 22 & 37 & 29 & 16 & 58 & 54 & 22 & 34.00 \\
\cmidrule(lr){2-10}
& FlexMoE
& \stat{31.33}{1.53}
& \stat{61.33}{2.31}
& \stat{46.33}{0.58}
& \stat{23.00}{1.00}
& \stat{73.33}{0.58}
& \stat{65.00}{2.65}
& \stat{26.00}{1.00}
& \stat{46.62}{0.81} \\
& Ours (CP-FT) & 33 & 64 & 46 & 28 & 73 & 62 & 27 & 47.57 \\
& FlexMoE (SP-XFER)
& \stat{34.33}{2.08}
& \stat{64.33}{3.05}
& \stat{50.33}{1.15}
& \stat{23.33}{0.58}
& \stat{73.67}{1.53}
& \stat{70.33}{0.58}
& \stat{26.33}{1.53}
& \stat{48.95}{0.84} \\

\midrule
\multicolumn{10}{c}{\textbf{Phi-3.5-MoE} (16 experts per layer, top-2 activated per token)} \\
\midrule

0\%
& Base model
& 56 & 77 & 68 & 40 & 79 & 76 & 38 & 62.00 \\

\midrule
20\%
& MoE-SVD$^{\dagger}$
& 54 & 81 & 65 & 39 & 79 & 74 & 36 & 61.14 \\
& MoE-SVD$^{*}$
& 55 & 78 & 66 & 38 & 80 & 72 & 35 & 60.57 \\
& TD-MoE$^{\dagger}$
& 55 & 77 & 65 & 39 & 79 & 74 & 38 & 61.00 \\
& TD-MoE$^{*}$
& 52 & 76 & 63 & 35 & 78 & 71 & 34 & 58.43 \\
\cmidrule(lr){2-10}
& FlexMoE
& \stat{52.33}{1.53}
& \stat{78.33}{0.58}
& \stat{64.67}{0.58}
& \stat{39.00}{1.73}
& \stat{76.33}{1.53}
& \stat{78.00}{1.00}
& \stat{33.00}{1.00}
& \stat{60.24}{0.36} \\

& FlexMoE (SP-XFER)
& \stat{56.00}{1.00}
& \stat{83.00}{0.00}
& \stat{65.33}{0.58}
& \stat{39.67}{1.15}
& \stat{79.00}{1.00}
& \stat{80.67}{0.58}
& \stat{39.33}{0.58}
& \stat{63.29}{0.00} \\

\midrule
40\%
& NAEE$^{\dagger}$
& 48 & 73 & 61 & 35 & 76 & 73 & 37 & 57.57 \\
& MoE-I$^2$$^{\dagger}$
& 40 & 59 & 27 & 29 & 70 & 67 & 25 & 45.29 \\
& MoE-SVD$^{\dagger}$
& 48 & 72 & 58 & 35 & 75 & 72 & 31 & 55.86 \\
& MoE-SVD$^{*}$
& 48 & 71 & 63 & 33 & 76 & 70 & 28 & 55.57 \\
& TD-MoE$^{\dagger}$
& 50 & 75 & 61 & 35 & 78 & 73 & 33 & 57.86 \\
& TD-MoE$^{*}$
& 47 & 73 & 60 & 30 & 76 & 69 & 31 & 55.14 \\
\cmidrule(lr){2-10}
& FlexMoE
& \stat{50.00}{1.73}
& \stat{74.67}{2.08}
& \stat{58.00}{0.00}
& \stat{35.00}{1.00}
& \stat{76.00}{0.00}
& \stat{65.33}{1.53}
& \stat{29.00}{1.00}
& \stat{55.43}{0.43} \\
& FlexMoE (SP-FT)
& \stat{53.00}{0.00}
& \stat{80.67}{1.15}
& \stat{59.00}{1.00}
& \stat{39.00}{1.00}
& \stat{78.33}{0.58}
& \stat{71.00}{1.00}
& \stat{36.33}{1.15}
& \stat{59.62}{0.33} \\

\midrule
60\%
& MoE-SVD$^{\dagger}$
& 40 & 60 & 46 & 30 & 71 & 68 & 25 & 48.57 \\
& MoE-SVD$^{*}$
& 42 & 63 & 48 & 31 & 74 & 68 & 24 & 50.00 \\
& TD-MoE$^{\dagger}$
& 41 & 70 & 46 & 28 & 73 & 68 & 23 & 49.86 \\
& TD-MoE$^{*}$
& 36 & 65 & 43 & 26 & 68 & 64 & 21 & 46.14 \\
\cmidrule(lr){2-10}
& FlexMoE
& \stat{43.67}{1.53}
& \stat{71.00}{1.00}
& \stat{47.33}{0.58}
& \stat{29.00}{1.00}
& \stat{73.33}{0.58}
& \stat{63.33}{0.58}
& \stat{25.33}{0.58}
& \stat{50.43}{0.25} \\
& FlexMoE (SP-XFER)
& \stat{46.00}{1.00}
& \stat{74.33}{0.58}
& \stat{51.33}{0.58}
& \stat{31.00}{1.00}
& \stat{75.00}{1.00}
& \stat{64.33}{0.58}
& \stat{27.67}{0.58}
& \stat{52.81}{0.22} \\

\midrule
\multicolumn{10}{c}{\textbf{Qwen2-57B-A14B} (8 shared + 64 routed experts per layer, 8 shared + top-8 routed activated per token)} \\
\midrule

0\%
& Base model
& 47 & 75 & 63 & 33 & 80 & 74 & 39 & 58.71 \\

\midrule
20\%
& NAEE$^{\dagger}$
& 42 & 72 & 59 & 31 & 79 & 72 & 36 & 55.86 \\
& MoE-SVD$^{\dagger}$
& 45 & 74 & 61 & 30 & 78 & 73 & 35 & 56.57 \\
& MoE-SVD$^{*}$
& 46 & 75 & 62 & 31 & 79 & 74 & 42 & 58.43 \\
& TD-MoE$^{\dagger}$
& 49 & 79 & 59 & 31 & 80 & 73 & 37 & 58.29 \\
& TD-MoE$^{*}$
& 47 & 78 & 59 & 33 & 80 & 73 & 36 & 58.00 \\
\cmidrule(lr){2-10}
& FlexMoE
& \stat{47.33}{0.58}
& \stat{75.33}{0.58}
& \stat{62.67}{0.58}
& \stat{33.67}{0.58}
& \stat{80.67}{0.58}
& \stat{74.33}{0.58}
& \stat{37.67}{0.58}
& \stat{58.81}{0.16} \\
& FlexMoE (SP-XFER)
& \stat{48.67}{0.58}
& \stat{78.67}{1.53}
& \stat{64.00}{1.00}
& \stat{34.33}{0.58}
& \stat{80.00}{1.00}
& \stat{75.33}{0.58}
& \stat{40.67}{0.58}
& \stat{60.24}{0.7} \\

\midrule
40\%
& NAEE$^{\dagger}$
& 39 & 72 & 54 & 28 & 76 & 71 & 32 & 53.14 \\
& MoE-SVD$^{\dagger}$
& 33 & 63 & 45 & 29 & 71 & 65 & 31 & 48.14 \\
& MoE-SVD$^{*}$
& 39 & 68 & 56 & 32 & 76 & 72 & 41 & 54.86 \\
& TD-MoE$^{\dagger}$
& 45 & 77 & 54 & 29 & 78 & 71 & 35 & 55.57 \\
& TD-MoE$^{*}$
& 48 & 77 & 50 & 29 & 77 & 72 & 35 & 55.43 \\
\cmidrule(lr){2-10}
& FlexMoE
& \stat{48.00}{0.00}
& \stat{76.67}{0.58}
& \stat{62.33}{0.58}
& \stat{33.00}{1.00}
& \stat{80.00}{0.00}
& \stat{73.67}{0.58}
& \stat{38.33}{1.15}
& \stat{58.86}{0.14} \\
& FlexMoE (SP-FT)
& \stat{49.67}{0.58}
& \stat{79.33}{1.15}
& \stat{62.00}{1.00}
& \stat{34.00}{1.00}
& \stat{79.67}{0.58}
& \stat{74.67}{0.58}
& \stat{40.67}{0.58}
& \stat{60.00}{0.42} \\

\midrule
50\%
& FlexMoE
& 47 & 76 & 61 & 33 & 80 & 73 & 40 & 58.57 \\

\midrule
60\%
& NAEE$^{\dagger}$
& 29 & 58 & 44 & 21 & 69 & 61 & 26 & 44.00 \\
& MoE-SVD$^{\dagger}$
& 32 & 62 & 44 & 27 & 69 & 64 & 30 & 46.86 \\
& MoE-SVD$^{*}$
& 42 & 71 & 46 & 29 & 74 & 68 & 32 & 51.71 \\
& TD-MoE$^{\dagger}$
& 41 & 73 & 46 & 28 & 74 & 68 & 31 & 51.57 \\
& TD-MoE$^{*}$
& 42 & 73 & 45 & 28 & 75 & 67 & 30 & 51.43 \\
\cmidrule(lr){2-10}
& FlexMoE
& \stat{49.00}{1.00}
& \stat{79.33}{0.58}
& \stat{59.33}{0.58}
& \stat{30.67}{0.58}
& \stat{80.00}{0.00}
& \stat{73.67}{1.15}
& \stat{38.33}{0.58}
& \stat{58.62}{0.08} \\
& FlexMoE (SP-XFER)
& \stat{50.67}{0.58}
& \stat{80.33}{1.53}
& \stat{59.67}{0.58}
& \stat{32.33}{1.15}
& \stat{79.00}{1.00}
& \stat{76.67}{0.58}
& \stat{39.67}{0.58}
& \stat{59.76}{0.58} \\

\midrule
80\%
& FlexMoE
& 44 & 73 & 54 & 29 & 77 & 73 & 32 & 54.57 \\

\midrule
\multicolumn{10}{c}{\textbf{Qwen3-30B-A3B (128 routed experts) (128 experts per layer, top-8 activated per token)}} \\
\midrule

0\%
& Base model
& 53 & 79 & 60 & 34 & 79 & 71 & 58 & 62.00 \\

\midrule
20\%
& FlexMoE (SP-XFER)
& 55 & 81 & 60 & 35 & 80 & 73 & 59 & 63.29 \\

\midrule
25\%
& FlexMoE
& 51 & 76 & 58 & 36 & 79 & 69 & 53 & 60.29 \\
& FlexMoE (SP-XFER)
& 54 & 81 & 59 & 35 & 80 & 74 & 58 & 63.00 \\

\midrule
40\%
& FlexMoE
& 50 & 74 & 54 & 33 & 77 & 65 & 49 & 57.43 \\
& FlexMoE (SP-FT)
& 51 & 81 & 57 & 34 & 79 & 70 & 53 & 60.71 \\

\midrule
60\%
& FlexMoE
& 36 & 66 & 43 & 29 & 72 & 60 & 35 & 48.71 \\
& FlexMoE (SP-XFER)
& 41 & 76 & 47 & 29 & 76 & 65 & 40 & 53.43 \\

\end{longtable}

\endgroup

\paragraph{Baseline reproduction and comparability.}
The source-reported baseline results in Table~\ref{tab:full-results} are retained only as
contextual references, since their calibration and recovery settings are not fully controlled.
In particular, the published methods are all data-dependent but use different calibration
corpora: NAEE and MoE-I$^2$ use C4-based calibration, whereas MoE-SVD and TD-MoE
use substantially smaller WikiText-2 calibration sets. To provide a fair
comparison, we locally reproduce the two strongest and latest baselines,
MoE-SVD and TD-MoE, using the same Zyda-2 calibration corpus, hardware environment,
pruning-budget accounting, and lm-eval-harness protocol as FlexMoE. We therefore base the
quantitative comparisons in the main text on these locally reproduced results ($^{*}$),
while retaining the published values here for completeness. 

During baseline reproduction, here are two factors that can make source-reported and locally reproduced results differ. First, decomposition-based methods estimate their
compression factors from a relatively small calibration subset and optimize a separate
compressed endpoint for each target budget. Their resulting decomposition behavior can therefore
vary with the calibration samples and random seed, making cross-corpus values not
directly interchangeable with our unified Zyda-2 reproduction. Second, for TD-MoE, we
found an additional difference in pruning-ratio accounting: its reported ratio is defined over
the selected MoE layers to which decomposition is applied, whereas our main comparison
uses the globally removed routed-expert parameters as the pruning ratio. The same nominal
TD-MoE ratio can therefore correspond to a smaller actual global parameter reduction.
For the other compared baselines, we verified that the parameter-ratio accounting is
consistent with our global routed-expert definition. This is another reason why our main
claims rely on the locally reproduced MoE-SVD and TD-MoE results. Notably, even against
the source-reported TD-MoE numbers under this more favorable compression ratio convention, the
shared-recovered FlexMoE model remains competitive or better across the matched budgets.

\paragraph{Additional validation on Qwen3-30B-A3B.}
We additionally evaluate Qwen3-30B-A3B~\citep{yang2025qwen3}, a newer and smaller-checkpoint MoE backbone
with 128 routed experts. Compared with our three main backbones, it provides a substantially
more expert-rich and strongly sparser routing topology. Results show that FlexMoE still remains effective on this architecture, and recovery performed only at the 40\% budget transfers well to both looser
and tighter masks. Because Qwen3-30B-A3B is newer and directly comparable results for
NAEE, MoE-I$^2$, MoE-SVD, and TD-MoE are unavailable under our seven-task protocol,
we use it only as additional architectural evidence to show backbone robustness rather than include it in the main baseline comparison, and exclude it from rerun and main text results.

\paragraph{Dense coverage of the budget trajectory.}
The 20\%, 40\%, and 60\% points in the main text are only shared comparison anchors with baselines, rather
than the only operating points supported by FlexMoE. Since the global budget is induced by
independent expert-wise actions, a four-action expert space still yields a much finer global
budget resolution (depends on the sparsity of MoE layers). In practice, FlexMoE's one-for-all action optimization stage allows to obtain subnetworks across the entire budget trajectory from 100\% to ~20\% retained expert parameters, with a fine granularity of approximately 0.5\% . So action training traverses a dense sequence of global pruning
ratios and masks can be exported throughout this trajectory, while each exported mask stores only
one action per expert and can be applied to the same importance-ranked full model, avoiding
the need to rerun action learning or store another full checkpoint. The additional intermediate
results in Table~\ref{tab:full-results} further show that the learned budget family remains stable
between the three main comparison points.

\subsection{Long-Context Evaluation on LongBench-E}
\label{app:longbench}

To test whether intra-expert channel pruning causes a disproportionate degradation as
context length increases, we additionally evaluate FlexMoE on LongBench-E~\citep{bai2024longbench},
the length-balanced evaluation setting introduced with LongBench.
We evaluated all three main backbones over the
13 LongBench-E tasks. Table~\ref{tab:longbench_res} reports these preliminary bucket-level averages.

\begin{table*}[t]
\centering
\caption{LongBench-E results across different pruning ratios.
Scores are reported in percentage points and grouped by context length.}
\label{tab:longbench_res}
\tiny
\resizebox{\textwidth}{!}{
\begin{tabular}{lccccccccc}
\toprule
& \multicolumn{3}{c}{Qwen2-57B-A14B}
& \multicolumn{3}{c}{Mixtral-8×7B}
& \multicolumn{3}{c}{Phi-3.5-MoE } \\
\cmidrule(lr){2-4}
\cmidrule(lr){5-7}
\cmidrule(lr){8-10}
Prune ratio
& 0--4k & 4--8k & 8k+
& 0--4k & 4--8k & 8k+
& 0--4k & 4--8k & 8k+ \\
\midrule
Base
& 44.43 & 41.02 & 40.55
& 39.85 & 40.50 & 36.70
& 41.73 & 42.16 & 42.61 \\

20\%
& 44.43 & 41.12 & 39.68
& 35.43 & 37.31 & 34.41
& 37.32 & 38.96 & 39.45 \\

40\%
& 42.92 & 40.33 & 40.71
& 31.15 & 30.71 & 29.23
& 36.70 & 36.16 & 36.74 \\

60\%
& 36.38 & 33.69 & 32.34
& 22.39 & 24.24 & 23.84
& 28.32 & 28.04 & 28.04 \\
\bottomrule
\end{tabular}
}
\end{table*}

As expected, stronger pruning lowers absolute model quality.  However, across all models, the results do not show abrupt representation collapse in the longer 4–8k and 8k+ bucket. At a fixed pruning
budget, performance across 4--8k and 8k+ contexts generally remains comparable to the
shorter bucket or follows the same context-length trend already present in the corresponding original 
full model. These results suggest that FlexMoE primarily reduces overall model capacity
 as expected to enhance deployability, rather than disproportionately damaging long-context representations as sequence length increases.

\subsection{Action-Learning Robustness: Cost Pressure and Calibration Resampling}
\label{app:action-robustness}

\paragraph{Cost-pressure schedule.}
The cost coefficient $\lambda_{\mathrm{cost}}(t)$ controls how quickly action optimization traverses the pruning-budget axis. 
A stronger schedule pushes experts toward thinner actions earlier, so a target pruning ratio
is encountered after fewer optimization steps and fewer observed calibration samples.
This accelerates training, but leaves less opportunity for the quality-preservation objective
to distinguish which expert-width assignments should absorb the pruning budget.
Conversely, a gentler schedule takes more exploration actions, spends more updates and calibration samples at intermediate
budgets, produces more fine-grained budget trajectory, effectively increasing quality-guided exploration before reaching an aggressive
pruning point.

We verify this trade-off on Qwen2-57B-A14B by comparing two linear cost schedules:

\begin{table}[t]
\centering
\caption{
Sensitivity to the cost-pressure schedule on Qwen2-57B-A14B.
``Step'' is the action-learning step at which the corresponding global pruning budget is reached.
The gentler schedule requires more optimization but yields consistently stronger masks.
}
\label{tab:cost_schedule}
\small
\begin{tabular}{cccc}
\toprule
Prune budget & Cost schedule & Reached at step & Seven-task Avg. \\
\midrule
20\% & $1 \rightarrow 15$   & 220  & 57.57 \\
     & $0.1 \rightarrow 5$  & 822  & $\mathbf{58.81\pm0.16}$ \\
\midrule
40\% & $1 \rightarrow 15$   & 225  & 57.86 \\
     & $0.1 \rightarrow 5$  & 1114 & $\mathbf{58.86\pm0.14}$ \\
\midrule
60\% & $1 \rightarrow 15$   & 266  & 54.86 \\
     & $0.1 \rightarrow 5$  & 4500 & $\mathbf{58.62\pm0.08}$ \\
\bottomrule
\end{tabular}
\end{table}

The gentler $0.1\!\rightarrow\!5$ schedule substantially improves pruning quality,
particularly at 60\%, where the optimizer receives substantially more quality-guided updates
before reaching the target budget. We therefore view $\lambda_{\mathrm{cost}}$ as a
trajectory-speed controller: stronger cost pressure trades search quality for a faster
budget traversal, while weaker pressure allocates more optimization and calibration samples
per unit of pruning. This also explains why the preferred scale of $\lambda_{\mathrm{cost}}$
is backbone-dependent: in a more strongly sparse architecture such as Qwen2, reducing the
width of one routed expert perturbs the full-model output less, so the same cost force can
move actions toward thin experts much faster.

\paragraph{Calibration resampling and seed robustness.}
The main experiments use the same selected cost schedule for each model, but rerun the complete FlexMoE
pipeline three times with different random seeds. Each run independently resamples Zyda-2
calibration batches and therefore changes the channel-ranking permutation, Gumbel action
trajectory, and recovery optimization. Despite these changes, the resulting masks are stable. This indicates that FlexMoE's strong performance are not tied to one
specific calibration subset and action trajectory, but learns structured, inner-expert level importance representations under given cost pressure, thus achieving a balance between output quality and parameter pruning.

\subsection{Ablation: Importance-Aware Channel Ranking}
\label{app:ranking-ablation}

We ablate the importance-aware reordering stage on Mixtral-8x7B. The goal is to test whether prefix slicing remains effective without first sorting expert FFN channels by importance. We compare three variants: base model, which directly applies our action-learning pipeline on the original unranked model; 100 ranked, where channels are ranked using 100 calibration samples; and 5k ranked, where the same reordering procedure uses 5,000 calibration samples. A subtle but important point is that, at the same target prune ratio, the action masks are not shared across these variants. Instead, each model variant runs its own action-learning process and uses the exported mask from that run. This is the fairest comparison: different channel orderings change the meaning of prefix slicing itself, so forcing the same mask across different orderings would confound the effect of reordering with an incompatible pruning pattern. In contrast, our current setup keeps the overall training pipeline identical to each model and only changes the ranking stage to obtain model variants. For each model variant, the action masks at different prune ratios are all exported from a single action-training run. There's no recovery fine-tuning applied.

\begin{table}[H]
\centering
\small
\setlength{\tabcolsep}{4pt}
\renewcommand{\arraystretch}{1.05}
\caption{Ablation on importance-aware reordering for prefix slicing on Mixtral-8x7B. All numbers are zero-shot accuracy (\%). ``100 ranked'' and ``5k ranked'' denote that the channel ordering is estimated using 100 and 5,000 calibration samples, respectively.}
\label{tab:ablation_reorder}
\begin{tabular}{llcccccccc}
\toprule
Ratio & Method & ARC-c & ARC-e & HellaS & OBQA & PIQA & WinoG & MathQA & Avg \\
\midrule
20\% & base model & 50 & 79 & 59 & 24 & 79 & 56 & 34 & 54.43 \\
20\% & 100 ranked & 51 & 79 & 62 & 34 & 81 & 75 & 36 & 59.71 \\
20\% & 5k ranked  & 53 & 81 & 62 & 33 & 81 & 73 & 37 & \textbf{60.00} \\
\midrule
40\% & base model & 44 & 71 & 51 & 21 & 75 & 52 & 28 & 48.86 \\
40\% & 100 ranked & 41 & 72 & 55 & 29 & 76 & 71 & 30 & 53.43 \\
40\% & 5k ranked  & 39 & 70 & 56 & 30 & 78 & 72 & 32 & \textbf{53.86} \\
\bottomrule
\end{tabular}
\end{table}

Table~\ref{tab:ablation_reorder} shows that importance-aware reordering is crucial for making prefix slicing effective.  Without reordering, the unreordered base model reaches only 54.43 average accuracy at the 20\% prune ratio and 48.86 at 40\%. In contrast, the ranked variants substantially improve performance, reaching 59.71/60.00 at 20\% and 53.43/53.86 at 40\%. The improvement is especially clear at the tighter 40\% budget, where preserving a more informative prefix matters more. This supports our core motivation: prefix channel retention action is only meaningful when the channel layout has first been transformed into an importance-ordered space. Once a modest number of calibration samples (steps) is available, the learned order converges and already becomes stable enough to support effective action learning, and further increasing the ranking set yields only marginal gains. This suggests that the ranking stage is not only effective but also practically sample-efficient: the importance estimates converge quickly, and the resulting channel-ranked layout is robust enough to support top-retained channel prefix slicing actions.

\subsection{Ablation: Efficiency of Single Point Fine-Tuning (SP-FT) Recovery}
\label{app:ft-abletion}
\paragraph{Cross-budget Fine-Tuning (CP-FT) Ablation Details.}
\label{app:cpft-details}

To obtain one recovered model that can serve multiple action masks at different budgets, we first implemented a cross-point fine-tuning strategy (\textsc{CP-FT}) inspired by the sandwich rule in universally slimmable networks (US-Nets) and the many-subnet distillation idea of AmoebaLLM \cite{yu2019usnets, fu2024amoeballm}. In each micro-step, we load a pool of trained action masks across budgets. We attached LoRA to full expert FFN projections and activate it for both the full teacher and subnet forwards. The full subnet is constrained with standard language-model CE loss to prevent full model degradation and affecting the inplace distillation objectives of subnets. While each sampled subnet is trained by soft distillation against the full-subnet teacher. Following the AmoebaLLM-style balancing factor used in our implementation, each subnet distillation loss is reweighted by the ratio between the magnitude of the full-subnet CE loss and that of the subnet distillation loss, so that no low-budget subnets dominates optimization because of higher distillation loss. Concretely, with full-subnet loss $\mathcal{L}_{\mathrm{full}}$ and sampled subnet losses $\{\mathcal{L}_{\mathrm{sub}}^{(i)}\}_{i=1}^{K}$, we optimize
\begin{equation}
\mathcal{L}_{\mathrm{CP\text{-}FT}}
=
\mathcal{L}_{\mathrm{full}}
+
\sum_{i=1}^{K}
\frac{\left|\mathcal{L}_{\mathrm{full}}\right|}{\left|\mathcal{L}_{\mathrm{sub}}^{(i)}\right|+\epsilon}
\,
\mathcal{L}_{\mathrm{sub}}^{(i)},
\end{equation}
where $\mathcal{L}_{\mathrm{full}}$ is the full-subnet CE loss and each $\mathcal{L}_{\mathrm{sub}}^{(i)}$ is the distillation loss from a sampled action mask. For each optimization step, the sampled masks (subnets) follow a US-Nets sandwich pattern consisting of the minimum-ratio mask and several random intermediate masks from the action pool, while the full model is always included as teacher.

\begin{figure*}[t]
\centering
\begin{minipage}[t]{0.5\textwidth}

\vspace{0pt}
\centering
\includegraphics[width=\linewidth]{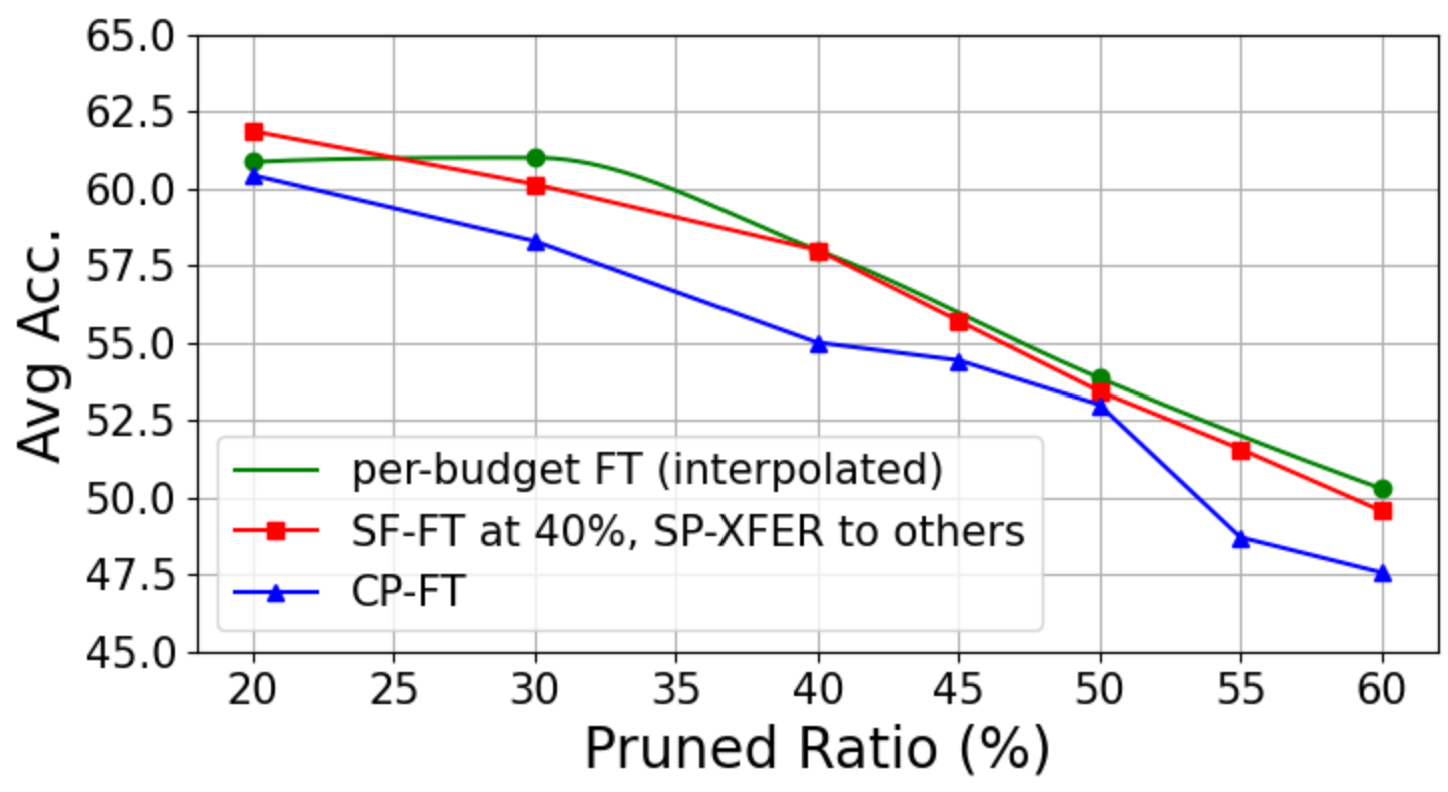}
\captionsetup{type=figure}
\vspace{-0.6em}\caption{Average downstream accuracy of different recovery strategies on Mixtral-8x7B across prune budgets.  Red curve is the recovery strategy used in FlexMoE}
\label{fig:ft_method_compare}

\end{minipage}\hfill
\begin{minipage}[t]{0.46\textwidth}
\vspace{0pt}
\centering
\includegraphics[width=\linewidth]{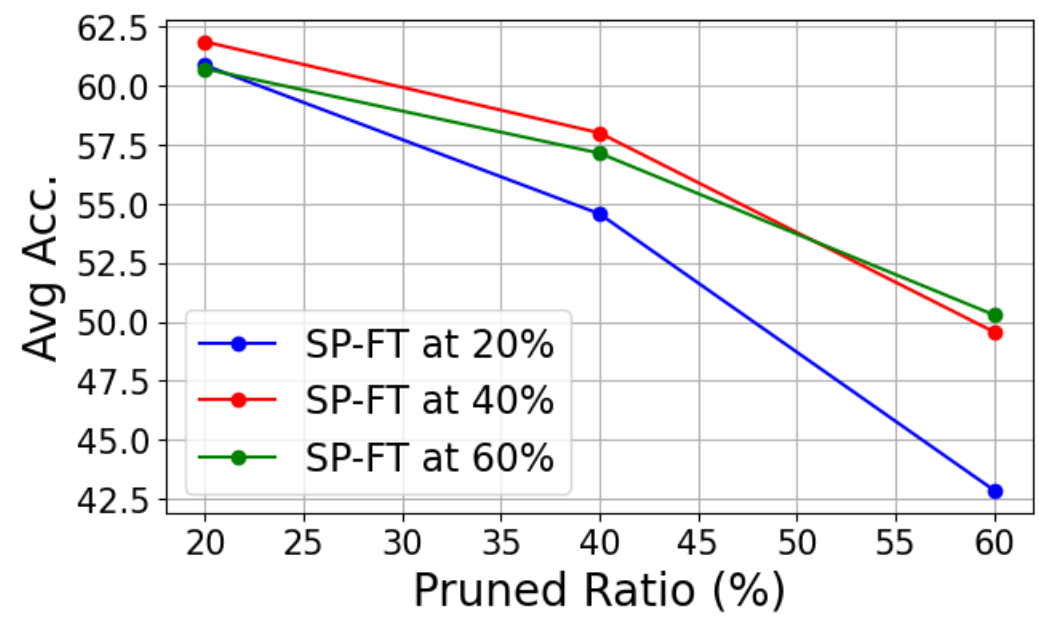}
\captionsetup{type=figure}
\vspace{-0.8em}\caption{Recovery-point generalization curve, plot by results shown in Table~\ref{tab:spft_cross_budget_transfer}. Red curve is the recovery point used in FlexMoE}
\label{fig:ft_point_ablation_curve}
\end{minipage}
\end{figure*}

\paragraph{Why Single Point Fine Tuning (\textsc{SP-FT}) is More Effective in Practice.}
We plot ablation results in Figure~\ref{fig:ft_method_compare}, with our SP-FT approach and per-budget fine-tuning results, the accuracy values are averaged from same 7 tasks in main experiments. We also found that this \textsc{CP-FT} strategy underperforms both per-budget fine-tuning and \textsc{SP-FT} over most budgets, whereas \textsc{SP-FT} followed by direct transfer already stays much closer to the per-budget fine-tuning results. We believe the reason is structural. \textsc{CP-FT} jointly optimizes many masks with substantially different active channel prefixes, so the shared LoRA update must satisfy several incompatible subnet configurations at once. In practice this might still weaken specialization and blur the recovery signal at any one budget. By contrast, leveraging the invariant nesting property of frozen route and expert topology, \textsc{SP-FT} concentrates all recovery capacity on one concrete subnet and then reuses the merged weights across nearby masks. Although this sacrifices a small amount of accuracy relative to independently fine-tuning every budget, it removes the need to train, store, and maintain one recovered model per budget point.

\paragraph{Why the Middle Budget Generalizes Best.}
 Table~\ref{tab:spft_cross_budget_transfer} and Figure~\ref{fig:ft_point_ablation_curve} show a clear asymmetry. Fine-tuning at the high-budget point (20\%) generalizes poorly when transferred downward: performance degrades monotonically as pruning becomes stronger, which is consistent with the intuition that high-budget recovery still spends its capacity on tail channels that are later removed by tighter masks. Interestingly, it is also not the best even on the 20\% target itself, where the mid-budget recovery point (40\%) performs better. A plausible explanation is that moderate masking might act as a useful regularizer: compared with 20\%-FT, 40\%-FT is forced to recover only the more reusable and important prefix channels, which improves robustness even at slightly looser budgets. On the other side, low-budget recovery (60\%) behaves as expected: it is stronger near its own fine-tuned budget, but its upward transfer remains weaker because the recovered update only covers a relatively small active prefix and therefore cannot adequately restore the additional channels exposed at looser masks. Overall, the 40\% recovery point provides the best global trade-off between coverage and specialization: it does not fully optimize any endpoint, but it yields the strongest one-step generalization across the whole budget family.

\section{Kernel Co-Design Implementation Details}
\label{app:kernel-details}

Our deployment goal is a single shared, importance-ordered MoE checkpoint that
stays fully resident and serves every pruning budget, with the serving
budget switched at runtime (Section~\ref{sec:deploy}). This imposes two
requirements on the expert forward. \textbf{(R1)} A smaller per-expert retained
width $k_e$ must translate into proportionally less GPU work, not merely a masked
full-width computation. \textbf{(R2)} The entire routed-expert forward must be
capturable in a single CUDA graph: In decoding the per-step cost is otherwise
dominated by kernel launch and host-side scheduling, which does not shrink
with $k_e$, so without graph capture the pruning savings stay hidden behind launch
overhead. 

\subsection{Naive Python Online Clipped FFN}

Our first online clipping approach was directly implemented based on HuggingFace-style original model forward path and applies budget-conditioned weights clipping inside the expert FFN forward at runtime. This path first identifies all routed experts in the current batch, and then iterates over them. In the original full-path (without weights pruning) implementation, the gate and up projections are stored contiguously as a gate--up weight tensor, so one expert forward can be executed with a single larger GEMM to enable larger GPU utilization, followed by a chunk into gate and up projection outputs. By contrast, online clipping breaks this fast path: once a runtime mask requests a retained width $k_e<I$, the implementation must first slice gate rows and up rows separately, concatenates into a clipped contiguous gate--up weight tensor, slice the down projection to the same width, and then launch the matrix multiplication and chunk. Since this happens inside a Python-side per-expert loop, the runtime pays both dispatch overhead and extra tensor-manipulation overhead before the actual GEMM.

\setlength{\algomargin}{1.5em}
\begin{algorithm}[t]
\caption{Naive Python Online Clipped FFN}
\label{alg:naive_python_online}

\KwIn{hidden states $X$, routed expert assignments, router weights, per-expert retained widths $\{k_e\}$, dense expert weights $\{W^{gu}_e, W^{down}_e\}$}
\KwOut{final routed experts FFN output $Y$}

Initialize $Y \leftarrow 0$\;
Find all active experts in the current batch\;
\ForEach{active expert $e$}{
    Gather routed token states $X_e$ and router weights for expert $e$\;
    Read retained width $k_e$\;
    \eIf{$k_e = I$}{
        \tcp{Original full path: use full contiguous gate-up weight}
        $Z^{gu}_e \leftarrow X_e (W^{gu}_e)^\top$\;
        $(G_e, U_e) \leftarrow \operatorname{chunk}(Z^{gu}_e, 2)$\;
        $H_e \leftarrow \operatorname{SiLU}(G_e)\odot U_e$\;
        $O_e \leftarrow H_e (W^{down}_e)^\top$\;
    }{
        \tcp{Online clipped path: extra slicing and reconstruction}
        $W^{gate}_e \leftarrow W^{gu}_e[0\!:\!k_e,:]$\; \tcp*[r]{extra slice}
        $W^{up}_e \leftarrow W^{gu}_e[I\!:\!I+k_e,:]$\; \tcp*[r]{extra slice}
        $\widehat W^{down}_e \leftarrow W^{down}_e[:,0\!:\!k_e]$ \tcp*[r]{extra slice}
        $\widehat W^{gu}_e \leftarrow \operatorname{cat}(W^{gate}_e, W^{up}_e)$ \tcp*[r]{hotspot}
        $\widehat Z^{gu}_e \leftarrow X_e (\widehat W^{gu}_e)^\top$\;
        $(\widehat G_e,\widehat U_e) \leftarrow \operatorname{chunk}(\widehat Z^{gu}_e,2)$\;
        $\widehat H_e \leftarrow \operatorname{SiLU}(\widehat G_e)\odot \widehat U_e$\;
        $O_e \leftarrow \widehat H_e (\widehat W^{down}_e)^\top$\;
    }
    Weight $O_e$ by router scores and scatter-add into $Y$\;
}
\Return{$Y$}
\end{algorithm}

This implementation is simple but inefficient for two reasons. First, once the learned action mask assigns different retained widths to different experts, the runtime can no longer naturally batch expert FFNs into one regular MoE matrix multiplication; it instead degrades toward many small expert-wise GEMMs plus Python-side scheduling. Second, compared with the dense full path, clipped execution inserts extra slice and cat operations on the weight tensors before every clipped expert forward. Especially for tensor concatenation step, where PyTorch's internal implementation involves additional empty memory allocation and tensor data copying. These overheads are specific to our nested prefix-slicing setting: they arise because the scenario is asked to materialize budget-specific clipped subnetworks on the fly from one shared MoE checkpoint across all budget family.

\subsection{Online-Reordered Shared Weight Layout}

To reduce the online clipping overhead, we first arrange the cross-budget shared model weights into a layout that is more convenient for runtime prefix slicing actions. In the original expert, the connected gate-up projection weight is stored as
\[
W^{gu}_{\mathrm{orig}}=
[g_0,\ldots,g_{I-1},u_0,\ldots,u_{I-1}],
\]
so to obtain $W^{gu}_{\mathrm{clip}}$ with only top $k$ channels requires two separate row slices followed by one concatenation:
\[
W^{gate}_{[:k]},\; W^{up}_{[:k]},\; W^{gu}_{\mathrm{clip}}= \operatorname{cat}(W^{gate}_{[:k]},W^{up}_{[:k]})
\]

Meanwhile, the original down projection is stored as $W^{down}_{\mathrm{orig}}\in\mathbb{R}^{H\times I}$, so clipping also requires a larger overhead column slice.

Our export path rewrites the shared weights into this new weights layout:
\[
W^{gu}_{\mathrm{reord}}=
[g_0,u_0,g_1,u_1,\ldots,g_{I-1},u_{I-1}],
\qquad
W^{down}_{\mathrm{reord}}=(W^{down}_{\mathrm{orig}})^\top \in\mathbb{R}^{I\times H}.
\]
Under this layout, clipping to $ W^{gu}_{\mathrm{clip}}$ with top $k$ channels becomes one prefix slice on the interleaved gate-up tensor and one row slice on the transposed down tensor:
\[
W^{gu}_{\mathrm{clip}} = W^{gu}_{\mathrm{reord}}[0:2k,:],
\qquad
W^{down}_{\mathrm{clip}} = W^{down}_{\mathrm{reord}}[0:k,:].
\]

In the original online path, the runtime must rebuild a $W^{gu}_{\mathrm{clip}}$ before every clipped expert forward. In the reordered layout, we can obtain requested weights by performing only 1 direct prefix slicing. This reduces the online scheduling cost from repeated tensor assembly.

\subsection{Graph-Capturable Fused Clipped FFN Forward}
\label{app:dkernel}

On top of the reordered shared weights, we co-design a fused Triton kernel that removes every host-side, input-dependent operation for decoding, so the whole routed-expert forward is captured once and
replayed. It consumes the same reordered layout as Section~B.2: the interleaved
gate--up tensor $\tilde W^{gu}\!\in\!\mathbb{R}^{E\times 2I\times H}$ (gate channel
$j$ at row $2j$, up channel $j$ at $2j{+}1$) and the transposed down tensor
$\tilde W^{down}\!\in\!\mathbb{R}^{E\times I\times H}$, so retaining the top-$k_e$
mid channels reduced to just one prefix in the channel dimension. Algorithm~\ref{alg:dkernel}
summarizes the forward; three design choices make it both budget-proportional (R1)
and capturable (R2).

\textbf{Device-resident fixed-shape routing.}
Routing is turned into fixed-size, device-resident block metadata by a single
device kernel (\textsc{MoEAlignBlockSize} from SGLang): a padded \texttt{sorted\_ids}
list, a per-block \texttt{expert\_ids} array, and a \texttt{num\_post} count. No
\texttt{bincount}, no device-to-host sync, and no host bucketing; across decode steps
only the \emph{values} of these buffers change, never their shapes. The token gather
is folded into the first GEMM (each block reads its hidden rows through
\texttt{sorted\_ids}) and the scatter is folded into the second GEMM's epilogue, so
there is no separate indexing step.

\textbf{On-device pruning under a fixed launch grid.}
Both GEMMs launch on a grid sized to the \emph{full} width
($\lceil I/B_N\rceil$ and $\lceil H/B_N\rceil$ column blocks), independent of the
budget. Each program reads its expert's retained width $k_e$ and realizes pruning
purely by data-dependent control flow on the device: in GEMM-1 a column block of mid
channels early-returns when its offset $n_0\!\ge\!k_e$ (the pruned channels are never
computed), and in GEMM-2 the contraction loop runs only over
$k_0\!\in\![0,k_e)$. The work thus shrinks with $k_e$ while the launch shape, kernel
arguments, and grid stay constant --- exactly what a CUDA graph requires.

\textbf{Width read from a device buffer at replay time.}
Crucially, $k_e$ is read from a device-resident buffer $\mathrm{keep}\!\in\!\mathbb{Z}^{E}$
(\texttt{keep[e]}) \emph{inside} the kernel, not baked into the launch configuration.
Because a captured graph replays its recorded launches but re-reads buffer
\emph{contents}, a single captured graph serves every budget: switching the serving
budget only overwrites $\mathrm{keep}$ (an $O(E)$ per layer device write), with no
graph re-capture and no weight reload (Section~\ref{sec:deploy}). To keep capture
deterministic the block sizes $B_M,B_N,B_K$ are fixed rather than autotuned.

Once replayed from a graph, decoding becomes GPU-bound, so the expert
weight-bandwidth term --- which \emph{does} scale with $k_e$ --- becomes a real
fraction of the step, and the pruning budget turns into the decode throughput
reported in Table~\ref{tab:online_kernel_throughput}.

\setlength{\algomargin}{1.5em}
\begin{algorithm}[t]
\DontPrintSemicolon
\SetKwInOut{Input}{Input}\SetKwInOut{Output}{Output}
\SetKwFunction{Align}{MoEAlignBlockSize}\SetKwFunction{Sum}{Sum}
\Input{hidden states $X\!\in\!\mathbb{R}^{T\times H}$, routed expert ids $\mathrm{eid}$, router weights $\alpha$; reordered interleaved $\tilde W^{gu}\!\in\!\mathbb{R}^{E\times 2I\times H}$ and transposed $\tilde W^{down}\!\in\!\mathbb{R}^{E\times I\times H}$; \emph{device-resident} retained widths $\mathrm{keep}\!\in\!\mathbb{Z}^{E}$}
\Output{routed-expert FFN output $Y\!\in\!\mathbb{R}^{T\times H}$}
\BlankLine
$(\mathrm{sid},\mathrm{eblk},n_{\mathrm{post}}) \leftarrow \Align(\mathrm{eid}, B_M, E)$\tcp*{fixed-shape device metadata, no host sync}
$n_{\mathrm{valid}} \leftarrow T\cdot K$\;
\BlankLine
\tcp{GEMM-1 (gate--up $+$ SiLU): fixed grid $(\lceil n_{\max}/B_M\rceil,\ \lceil I/B_N\rceil)$}
\ForEach{block $(p_m,p_n)$ \textbf{in parallel}}{
  \lIf{$p_m B_M \ge n_{\mathrm{post}}$ \textbf{or} $\mathrm{eblk}[p_m]<0$}{\textbf{return}}
  $e \leftarrow \mathrm{eblk}[p_m]$;\ \ $k_e \leftarrow \mathrm{keep}[e]$;\ \ $n_0 \leftarrow p_n B_N$\;
  \lIf{$n_0 \ge k_e$}{\textbf{return}\tcp*[f]{pruned channel block: no work}}
  $\mathrm{rows} \leftarrow \mathrm{sid}[\text{block}]$;\ \ $t \leftarrow \lfloor \mathrm{rows}/K\rfloor$\tcp*{fused token gather}
  $a_g,a_u \leftarrow 0$\;
  \For{$k_0 = 0$ \KwTo $H$ \textbf{step} $B_K$}{
     $a_g \mathrel{+}= X[t,k_0{:}]\cdot \tilde W^{gu}[e,\,2n,\,k_0{:}]$;\ \ $a_u \mathrel{+}= X[t,k_0{:}]\cdot \tilde W^{gu}[e,\,2n{+}1,\,k_0{:}]$\;
  }
  $\mathrm{mid}[\text{block},\,n] \leftarrow \mathrm{SiLU}(a_g)\odot a_u$\tcp*{masked to $n<k_e$}
}
\BlankLine
\tcp{GEMM-2 (down): fixed grid $(\lceil n_{\max}/B_M\rceil,\ \lceil H/B_N\rceil)$}
\ForEach{block $(p_m,p_n)$ \textbf{in parallel}}{
  \lIf{$p_m B_M \ge n_{\mathrm{post}}$ \textbf{or} $\mathrm{eblk}[p_m]<0$}{\textbf{return}}
  $e \leftarrow \mathrm{eblk}[p_m]$;\ \ $k_e \leftarrow \mathrm{keep}[e]$;\ \ $o \leftarrow 0$\;
  \For{$k_0 = 0$ \KwTo $k_e$ \textbf{step} $B_K$}{
     $o \mathrel{+}= \mathrm{mid}[\text{block},k_0{:}]\cdot \tilde W^{down}[e,k_0{:},\,:]$\tcp*{K-loop clamped to $k_e$}
  }
  $o \leftarrow o \cdot \alpha[\mathrm{sid}]$\tcp*{fused router weight}
  scatter $o$ into $\mathrm{out}[\mathrm{sid},\,:]$\tcp*{flat $[T\!\cdot\!K,\,H]$ buffer}
}
$Y \leftarrow \Sum(\mathrm{out}.\mathrm{view}(T,K,H),\ \mathrm{axis}{=}1)$\tcp*{fixed-shape top-$k$ reduction}
\Return $Y$\;
\caption{Graph-Capturable Fused Clipped FFN Forward (deployed decode path)}
\label{alg:dkernel}
\end{algorithm}

\section{Reproducibility Details}
\label{app:reproducibility}

\subsection{Reproduction Protocol}
\label{app:reproduction-protocol}

\paragraph{Models and calibration data, and evaluation.}
We apply FlexMoE to Mixtral-8$\times$7B, Phi-3.5-MoE, Qwen2-57B-A14B and Qwen3-30B-A3B using their original pretrained checkpoints and routing structures.
All three post-training stages---channel ranking, action learning, and recovery fine-tuning---use only the train split of our Zyda-2 \texttt{sample-100BT} subset.

\paragraph{Pruning protocol.}
The pruning budget is the fraction of routed-expert FFN parameters removed; routers and other non-routed-expert modules are unchanged.
We use the discrete expert-retention action space
$\mathcal A=\{0.1,0.4,0.7,1.0\}$.
Channel ranking is performed once before action learning.
During action learning, all pretrained model weights are frozen and only the expert action logits $\alpha$ are optimized, ensuring that masks obtained at different points of the trajectory refer to the same ranked weight set.
Each exported mask therefore stores only one retention action per routed expert and can be directly reapplied to the shared ranked model.

\paragraph{Recovery protocol.}
For the main recovery experiments, we select the action mask at 40\% pruning and freeze the importance-ranked base weights.
LoRA parameters are only attached to channels exposed at this budget, and only LoRA parameters attached are optimized at this stage.
After optimization, LoRA parameters are merged into the ranked base weights; the resulting recovered weight set is reused directly with the 20\% and 60\% masks as SP-XFER without further fine-tuning.

\subsection{Complete Hyperparameter Configuration}
\label{app:hyperparameters}

Table~\ref{tab:flexmoe_hyperparameters} summarizes the complete configuration used for the three backbones shown in main text results.
Unless explicitly listed otherwise, model-specific settings are identical across backbones.

\newcolumntype{C}[1]{>{\centering\arraybackslash}m{#1}}

\begingroup
\scriptsize
\setlength{\tabcolsep}{2.5pt}
\renewcommand{\arraystretch}{1.10}

\begin{longtable}{
    @{}
    C{0.105\textwidth}
    C{0.185\textwidth}
    C{0.225\textwidth}
    C{0.19\textwidth}
    C{0.225\textwidth}
    @{}
}

\caption{
Complete FlexMoE reproduction configuration for the three main backbones.
``Same'' indicates the same setting as Mixtral-8$\times$7B.
}
\label{tab:flexmoe_hyperparameters}\\

\hline
\textbf{Stage} &
\textbf{Hyperparameter} &
\textbf{Mixtral-8$\times$7B} &
\textbf{Phi-3.5-MoE} &
\textbf{Qwen2-57B-A14B} \\
\hline
\endfirsthead

\multicolumn{5}{c}{
\tablename~\thetable\ -- continued
}\\
\hline
\textbf{Stage} &
\textbf{Hyperparameter} &
\textbf{Mixtral-8$\times$7B} &
\textbf{Phi-3.5-MoE} &
\textbf{Qwen2-57B-A14B} \\
\hline
\endhead

\hline
\multicolumn{5}{r}{Continued on next page} \\
\endfoot

\hline
\endlastfoot

\hline\hline
\multicolumn{5}{c}{
\textbf{Importance-Aware Channel Ranking}
}\\
\hline\hline

Ranking
& Dataset
& Zyda-2 \texttt{sample-100BT}, train split
& Same
& Same \\
\hline

& Steps
& 700
& Same
& Same \\
\hline

& Batch
& Batch size 8; sequence length 1,024
& Same
& Same \\
\hline

& Weight update
& None; gradients are used only to estimate channel saliency
& Same
& Same \\
\hline

\hline\hline
\multicolumn{5}{c}{
\textbf{Expert-Wise Action Learning}
}\\
\hline\hline

Action
& Dataset
& Zyda-2 \texttt{sample-100BT}, train split
& Same
& Same \\
\hline

& Trainable parameters
& Action logits $\alpha$ only; pretrained weights frozen
& Same
& Same \\
\hline

& Retention actions
& $\{0.1,0.4,0.7,1.0\}$; full-action initialization bias 0.05
& Same
& $\{0.1,0.4,0.7,1.0\}$; full-action initialization bias 0.3 \\
\hline

& Optimizer
& AdamW; LR $2\times10^{-4}$; $\beta=(0.9,0.99)$;
weight decay 0; grad. clip 1.0
& Same
& Same \\
\hline

& Maximum steps
& 20,000; one sampled action mask per batch
& Same
& Same \\
\hline

& Batch
& Batch size 4; grad. accumulation 4; sequence length 1,024
& Same
& Batch size 4; grad. accumulation 8; sequence length 1,024 \\
\hline

& Quality loss
& CE weight 1; KL weight 1; distillation temperature $T=2$
& Same
& Same \\
\hline

& Gumbel temperature $\tau$
& Piecewise linear
$4.0\!\rightarrow\!0.8\!\rightarrow\!0.5$;
milestones 2,000 / 8,000
& Same
& Same \\
\hline

& Cost weight $\lambda_{\mathrm{cost}}$
& Linear $1\!\rightarrow\!15$ over 10,000 steps
& Same
& Linear $0.1\!\rightarrow\!5$ over 10,000 steps \\
\hline

& Entropy weight $\beta$
& Linear $0.2\!\rightarrow\!0$ over 5,000 steps
& Same
& Same \\
\hline

& LR schedule
& 200-step warmup; decay to $0.1\times$ initial LR
& Same
& Same \\
\hline

& Mask export interval
& Every 1\% decrease in global retention
& Every 5\%
& Every 5\% \\
\hline

\hline\hline
\multicolumn{5}{c}{
\textbf{Single-Point Recovery Fine-Tuning}
}\\
\hline\hline

Recovery
& Dataset
& Zyda-2 \texttt{sample-100BT}, train split
& Same
& Same \\
\hline

& Recovery budget
& 40\% pruning
& Same
& Same \\
\hline

& Trainable parameters
& LoRA only; ranked base weights frozen
& Same
& Same \\
\hline

& Optimizer
& AdamW; LR $1\times10^{-4}$; $\beta=(0.9,0.99)$;
weight decay 0; grad. clip 1.0
& Same
& Same \\
\hline

& Training steps
& Maximum 8,000; selected checkpoint 7,800
& Maximum 8,000; selected checkpoint 3,001
& Maximum 6,000; selected checkpoint 4,501 \\
\hline

& Batch
& Batch size 2; grad. accumulation 4; sequence length 2,048
& Same
& Same \\
\hline

& LR schedule
& 100-step linear warmup; decay to $0.1\times$ initial LR
& Same
& Same \\
\hline

& LoRA
& Rank 64; $\alpha=128$; dropout 0; no bias
& Same
& Same \\
\hline

& LoRA targets
& \texttt{gate\_up\_proj} and \texttt{down\_proj}
in all 32 MoE layers
& Same
& \texttt{gate\_up\_proj} and \texttt{down\_proj}
in all 28 MoE layers \\
\hline

& Recovery loss
& CE weight 1; KL weight 1; distillation temperature $T=2$
& Same
& Same \\
\hline

\end{longtable}

\endgroup

\subsection{Independent Runs, Statistics, and Compute}
\label{app:reproduction-statistics}

\paragraph{Independent runs.}
For the reported repeated-run experiments, we execute the complete pipeline three times with different seeds, and we report the sample mean and sample standard deviation $s$. Each run independently changes LoRA parameter initialization (action logits initialization are not affected by the random seed, since they are initialized as pointing to the full expert, and push them to thinner actions during training) and independently samples Zyda-2 batches, therefore each run can produce a different optimization trajectory for channel permutation, action-learning and recovered model. Despite rerun variation stochastic optimization, FlexMoE consistently converges to strong solutions across multi-budget with low performance variance, demonstrating the robustness of the learned compression trajectory.

\paragraph{Compute.}
Importance ranking is performed on one NVIDIA H200, while action learning and recovery fine-tuning use two H200 GPUs.
The observed action-learning wall-clock times are approximately 48, 30, and 7 hours for Mixtral-8$\times$7B, Phi-3.5-MoE, and Qwen2-57B-A14B, respectively, to optimize the action trajectory once and produce fine-grained budget points from 0\% to ~70\% prune ratio. By comparison, our local reproductions of MoE-SVD and TD-MoE require several hours to produce an independent compressed model for each target budget in 20\%, 40\% and 60\%, whereas a single FlexMoE action-learning run amortizes its cost over the entire fine-grained budget trajectory, with additional operating points represented only by lightweight action masks rather than separately trained and stored checkpoints.

\clearpage
\end{document}